\documentclass[journal]{IEEEtran}
\usepackage[font=footnotesize]{caption}
\usepackage[font=footnotesize]{subcaption}
\usepackage{amssymb} 
\usepackage{amsmath} 
\usepackage{bm}

\usepackage{algorithm}
\usepackage{algorithmic}
\usepackage{booktabs} 
\usepackage{multirow}
\usepackage[table]{xcolor}
\usepackage{pgfplots}
\usepackage[hidelinks]{hyperref}

\definecolor{Light0}{rgb}{0.98, 0.95, 0.99}
\definecolor{Light1}{rgb}{0.98, 0.95, 0.90}
\definecolor{Light2}{rgb}{0.98, 0.98, 0.93}
\definecolor{Light3}{rgb}{0.98, 0.98, 1}

\newcommand{\trsp}{{\scriptscriptstyle\top}}

\begin{document}
\title{Ergodic Exploration using Tensor Train:\\Applications in Insertion Tasks}
%
%
%

\author{Suhan~Shetty$^{1,2}$,
        Jo\~ao~Silv\'erio$^{1}$,
        and~Sylvain~Calinon$^{1,2}$
	\thanks{$^{1}$ Idiap Research Institute, Martigny, Switzerland.}
	\thanks{$^{2}$ \'Ecole Polytechnique Fed\'erale de Lausanne (EPFL),  Switzerland.} 
	\thanks{E-mails: name.surname@idiap.ch}
	\thanks{This work was supported by the CoLLaboratE project (https://collaborate-project.eu), funded by the EU within H2020-DT-FOF-02-2018 under grant agreement 820767, and by the LEARN-REAL project (https://learn-real.eu), funded by the Swiss National Science Foundation (CHIST-ERA project).}
}


\maketitle

\begin{abstract}
In robotics, ergodic control extends the tracking principle by specifying a probability distribution over an area to cover instead of a trajectory to track. The original problem is formulated as a spectral multiscale coverage problem, typically requiring the spatial distribution to be decomposed as Fourier series. This approach does not scale well to control problems requiring exploration in search space of more than 2 dimensions. To address this issue, we propose the use of tensor trains, a recent low-rank tensor decomposition technique from the field of multilinear algebra. The proposed solution is efficient, both computationally and storage-wise, hence making it suitable for its online implementation in robotic systems. The approach is applied to a peg-in-hole insertion task requiring full 6D end-effector poses, implemented with a 7-axis Franka Emika Panda robot. In this experiment, ergodic exploration allows the task to be achieved without requiring the use of force/torque sensors.

\end{abstract}


\IEEEpeerreviewmaketitle

\section{Introduction}
\label{intro}
\IEEEPARstart{A}{utonomous} systems are often encountered with coverage tasks such as localization, tracking, and active learning. In such tasks, the agent might be required to explore a region of its state space, either due to the nature of the task at hand (e.g. surveillance) or due to uncertainties induced by sensory inaccuracies (e.g. peg-in-hole insertion). In such problems, the coverage task can be specified by a reference probability density function, which encodes the importance of exploration at any point of the state space. For such problems, a pattern-based coverage approach (e.g., a ``lawnmower-type'' strategy), as commonly used in low-dimensional state space, is not scalable, and hence not applicable to most of the applications encountered in practice \cite{multiscale}. Maximizing information gain, another popular approach to circumvent uncertainty, is not suitable for exploration since the coverage is likely to be concentrated in regions around information maxima disproportionately over the period of exploration \cite{miller}.

\begin{figure}[t!]
    \centering
     \includegraphics[width=0.9\linewidth]{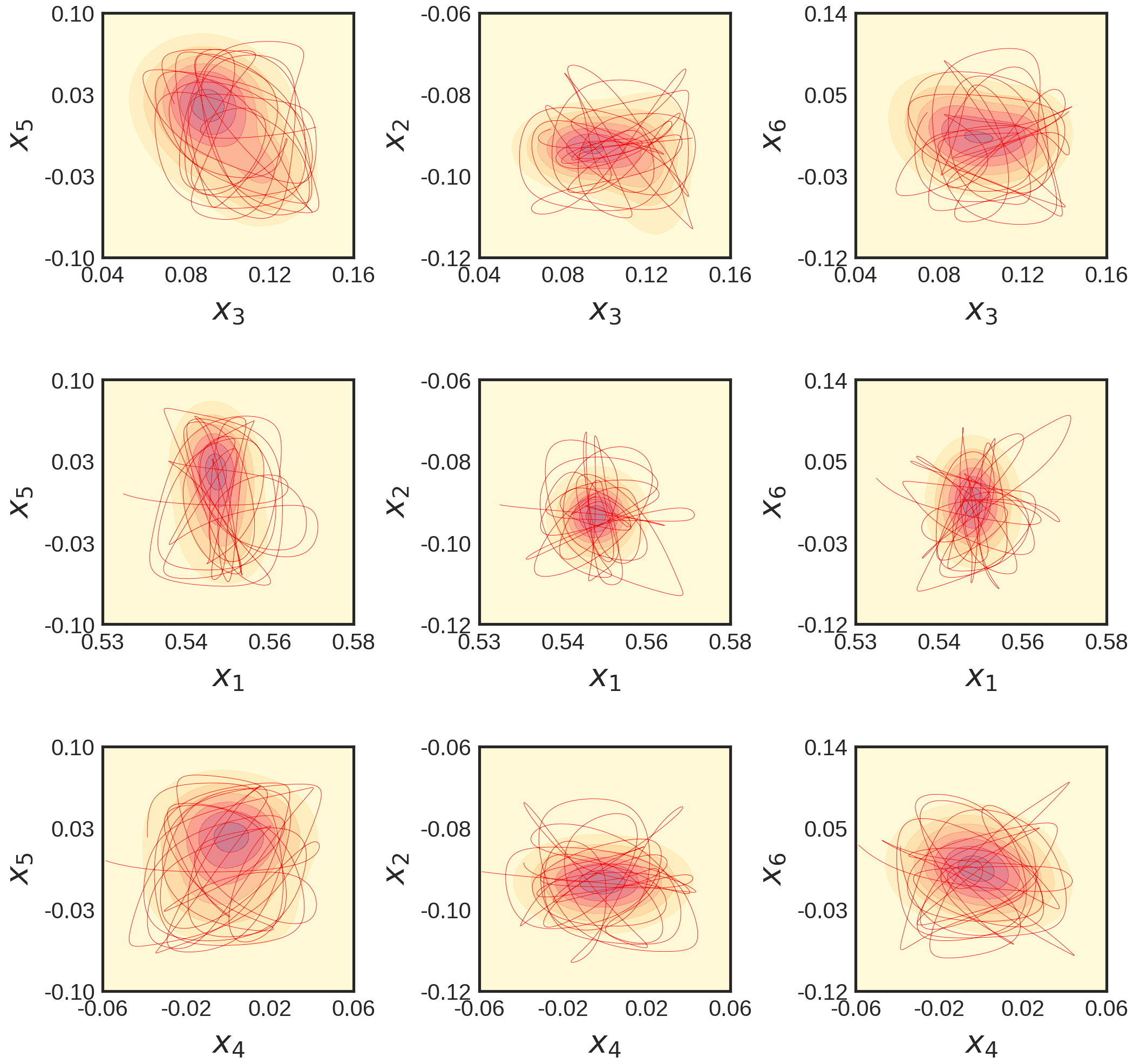}
     \caption{The 6D robot trajectory generated by ergodic controller offline for the insertion task.}
     \label{fig:erg_traj_robot_open_loop}
\end{figure}

Ergodic control provides an elegant solution to design control policies for such autonomous systems, in order to equip them with natural search behaviors. For a given reference probability density function over a domain of interest in the state space of the robot, a dynamical system is said to be ergodic if the fraction of time spent in a given region is proportional to the probability mass of that region \cite{mezic}. This is formalized in ergodic theory, where the goal is to characterize how ergodic a given dynamical system is. Ergodic control, on the other hand, aims to design a control policy for a given autonomous system so that the trajectory evolution of the resulting dynamical system is ergodic for the reference probability distribution. Systems engineered in such a way have already found applications in robotics \cite{miller, mathew_experimental}. The original approach to ergodic control is called Spectral Multiscale Coverage (SMC) and involves spectral analysis of the dynamical system evolution \cite{mezic}. This elegant method was proposed for point-mass systems having receding horizon control with infinitesimal control horizon. This original work paved the way for various extensions, with other types of dynamical systems and finite-horizon controllers \cite{mathew_experimental,miller,anastasia_erg}. The idea behind SMC \cite{mezic} is to minimize a metric, called the ergodic metric, that quantifies the match between the Fourier coefficients of a reference distribution and those of the time-averaged statistics of the system trajectory. As we will see in Section \ref{erg_define}, this method unfortunately suffers from the curse of dimensionality, prohibiting its applications to search spaces with more than 2 or 3 dimensions, which are often encountered in robot manipulation problems (e.g. exploration in the task space of an end-effector is often a 6D problem). 

In this paper, we propose a solution to overcome the challenges in SMC by using low-rank tensor approximation techniques, in the form of a tensor train (TT), and hence expanding the domain of ergodic control to robot manipulation. Figure \ref{fig:erg_traj_robot_open_loop} shows an ergodic exploration behavior generated by the proposed method for the peg-in-hole task considered in this paper.

We showcase our approach in a 6D peg-in-hole insertion task using a robot manipulator, by considering the position and orientation of the end-effector\footnote{A video of the experiment is available at \url{https://sites.google.com/view/ergodic-exploration/}}. In peg-in-hole scenarios, perception and modeling inaccuracies often compromise success, requiring the robot to leverage smart control strategies. Here, we propose to apply ergodic control to facilitate the insertion by letting the robot explore around the hole location in the 6D state space of the end-effector. In this application, we rely on human demonstrations to specify the distribution that the robot should use for an ergodic exploration. 

The main contribution of this work is an algorithm for SMC to generate ergodic exploration in multi-dimensional spaces, which was previously considered to be an intractable problem. In particular, we improve the state of the art by proposing:
\begin{itemize}
    \item Fast ways to compute the Fourier coefficients of multivariate functions, a well-known bottleneck in the ergodic control literature;

    \item The use of tensor train to exploit the inherent low-rank structure in the problem, which is used to overcome the curse of dimensionality in both real-time computation and storage requirements and facilitate implementation of ergodic control online on robotic systems.

\end{itemize}

The proposed ergodic control algorithm paves the way for two additional contributions in robotics. Particularly, we:
\begin{itemize}
    \item Extend ergodic control to peg-in-hole tasks solved with an online policy, with a novel, principled and theoretically-grounded exploratory strategy for insertion tasks that does not rely on specialized sensors but only on human demonstrations;
	\item Provide a formal way to perform ergodic exploration in orientation by relying on the $\mathcal{S}^3$ Riemannian manifold.
\end{itemize}
To the best of our knowledge, this is the first time ergodic control is implemented online on a physical robot for exploration in dimension greater than $2$. Note that the strategies to mitigate the curse of dimensionality introduced in this paper have the potential to be applied to many other applications in robotics to address real-time computation and limited storage requirements. Similarly, the proposed control strategy is not limited to manipulation applications, and can be extended to other robotics scenarios requiring high-dimensional coverage (e.g. 3D-object modeling, 6D surveillance).\footnote{The proposed algorithm has been numerically evaluated for state space of up to $15$ dimensions.} Finally, even though we rely on human demonstrations to obtain the reference distribution, this reference can in practice be specified/learned in different ways (e.g. from sensor uncertainty models).

The paper is organized as follows: Section \ref{survey} gives a literature survey on ergodic control, tensor methods and control strategies for insertion tasks. Section \ref{sec:problem} covers the necessary mathematical background to introduce our contribution. Particularly, Section \ref{tensors_define} introduces tensor algebra. In Section \ref{erg_define}, the mathematical formulation of ergodic control is described, where the underlying challenges of the algorithm are outlined. Section \ref{tensor_decomposition_define} briefly introduces tensor decomposition techniques and Section \ref{tt_define} gives an overview of the tensor train decomposition, which is the main tool for low-rank approximation used in this paper. In Section \ref{main_algo}, we propose low-rank approximation using the tensor train as a solution for multidimensional ergodic exploration. In Section \ref{numerical_evaluation}, we evaluate the proposed algorithm in simulation. Lastly, in Section \ref{experimental_evaluation} we showcase the results of our approach in a peg-in-hole insertion task using a torque-controlled 7-axis Franka Emika Panda robot.

\section{Motivation and Related Work}
\label{survey}
\subsection{Ergodic Control}
\label{erg_survey}
A solution to ergodic control was originally proposed by \cite{mezic} using Spectral Multiscale Coverage (SMC) in the form of a feedback control law  designed for multi-agent systems, with an objective defined so that the agents trajectories cover a reference probability distribution. Here, the system considered is a point-mass system. The control policy is obtained by solving an optimization problem with an ergodic metric as the cost function (see Section \ref{erg_define}). 
The ergodic metric compares the Fourier series coefficients associated to the spatial reference distribution and the trajectory evolution of the system.

Although other possible choices of basis functions would be interesting to investigate (e.g. wavelets), the Fourier transform holds essential properties that are relevant to the considered problem. In \cite{Calinon19MM}, we discuss the use of Fourier series within ergodic control, including their links to cosine basis functions, and their properties for reference distributions in the form of mixtures of Gaussians.

The ergodic metric can be used as a starting point to design other forms of controllers. For example, ergodic controllers have been proposed using nonlinear dynamical systems and finite control horizons \cite{miller}, using projection-based trajectory optimization \cite{miller_traj_opt}, or using hybrid systems theory  \cite{anastasia_erg}. An overview of these methods with finite control horizon can be found in \cite{ergodic_tutorial}. 

The main drawback of these methods is that they suffer from the curse of dimensionality (see Section \ref{erg_define}) when the dimension of the state space for ergodic exploration increases. This is due to the computational complexity and storage demanded in working with the ergodic metric and the control policy derived from it. For low dimensional exploration tasks (2D), Dressel and Kochenderfer used supervised learning to reduce the computational burden \cite{dressel_info_maps}. However, it does not scale to higher dimensional problems. Several authors deviated from the approach used in SMC to tackle this limitation. In \cite{kl_choset} and \cite{abraham_kle3}, the authors relied on a different ergodic metric based on a Kullback-Leibler (KL) divergence measure for finite sensor footprint, where the control policy is obtained using sampling-based techniques. Here, the ergodic metric (KL-divergence) is approximated using the samples from the reference distribution. Sampling-based methods avoid the curse of dimensionality but they can still be computationally expensive to address the real-time computational requirements of robotics systems. Moreover, the performance of the method is heavily dependent on the quality of the samples obtained, which is hard to assess. While most sampling-based methods generate the ergodic trajectory offline, its online implementation on real robots, which is the focus of the current paper, is still a challenging problem. Based on \cite{kl_choset}, Abraham \emph{et al.} provides an online version of ergodic control with KL-divergence as ergodic metric \cite{abraham_kle3}, at the expense of potentially losing ergodicity in the exploration (e.g., by limiting the search to high density regions). 

This paper keeps the original methodology (SMC) proposed by \cite{mezic}, which is the foundation of most literature on ergodic control, and which has the advantage of providing closed form solutions for many of the commonly used models of dynamical systems (kinematics-based) \cite{mathew_experimental}, while providing multi-scale coverage behaviour. To do so, we propose a solution based on a low-rank tensor approximation to overcome the curse of dimensionality. The proposed algorithm has intuitive hyperparameters that can be adjusted to address the storage and computational constraints of the application. 


\subsection{Tensor Methods}
\label{tensors_survey}
As we will see in Section \ref{erg_define}, the difficulty in ergodic control essentially arises from the storage and computation with multi-dimensional arrays involved in the algorithm. Tensor methods have recently gained popularity in signal processing, physics, applied mathematics, and machine learning communities due to their efficiency in storing and working with multidimensional arrays. These methods exploit the structure inherent in multidimensional arrays such as symmetry, parallel proportionality, and separability to represent them compactly and robustly. Furthermore, they allow performing algebraic operations efficiently in the compact format. Thus, the storage complexity and the algebraic operations complexity are significantly reduced. For a survey of classical tensor methods, we refer the readers to \cite{tensors_survey_kolda}. For applications of tensor methods in signal processing and machine learning, we refer to \cite{cichockiSP,tensors_survey_ml}. In control, tensor methods have been used in \cite{horowitz} and \cite{gorodetsky} to solve multidimensional optimal control problems which were previously considered to be intractable. 

\subsection{Insertion Tasks}
\label{experiments_survey}
Peg-in-hole insertion is a typical and important problem in robotics. Many strategies to solve this problem depend on expensive force and torque sensors \cite{force_feedback_insertion_1, force_feedback_insertion_2}. Sensorless strategies \cite{sensorless1, sensorless2, Ehlers19}, on the other hand, rely only on the state of the end-effector and provide a low cost solution. However, most of the sensorless strategies depend either on a predetermined trajectory that the robot end-effector needs to follow \cite{Ehlers19}, or a full modeling of the insertion behavior \cite{sensorless1, sensorless2}. In \cite{Ehlers19}, insertion is treated as a 3D (2D position and 1D orientation of the end-effector) trajectory tracking problem, where the reference trajectory for the robot end-effector is generated offline by using a coverage strategy such as ergodic control.
As we will show, this strategy fails for the peg-in-hole insertion task considered in this paper, as a trajectory generated offline is often not possible to track due to obstructions from the surface of the hole and the peg. To address this challenge, our approach instead formulates the coverage problem in an online manner, with exploration simultaneously in the full 6D state space of the robot end-effector (3D position and 3D orientation). 

In order to handle the exploration in orientation jointly with the position, we extend the control strategy to Riemannian manifolds by modeling the probability distribution of orientations, see \cite{Zeestraten17,Calinon20RAM} for details. Subsequently, we use an online implementation of ergodic control as the solution for coverage. The algorithm proposed in this paper allows us to run the ergodic controller online on the robot for 6D insertion tasks.

\begin{table}[!th]
	\caption{Description of key notations and variables.} 
	\resizebox{\linewidth}{!}{%
		{\renewcommand\arraystretch{1} 
			\centering
			\begin{tabular}{m{0.1\linewidth}m{0.65\linewidth}}
				\noalign{\hrule height 1.5pt}
				
				\rowcolor{Light0} $d$ & Dimension of the state space\\ 
				\rowcolor{Light0} $K$ & Number of elementary basis functions along each dimension ($K^d$ basis functions for the state space)\\ 
				\rowcolor{Light0} $N$ & Degree of Gaussian quadrature rule ($N\in \mathbb{Z}^{+}$)\\
				
				\rowcolor{Light1} $\bm{b}$ & 1D array (vector)\\
				\rowcolor{Light1} $\bm{A}$ & 2D array (matrix)\\
				\rowcolor{Light1} $\bm{\mathcal{G}}$ & tensor of order $>2$ \\
				\rowcolor{Light1} $\bm{\mathcal{G}_k}$ & $\bm{k}$-th element of a tensor $\bm{\mathcal{G}}$ (a scalar) \\
				\rowcolor{Light1} $\bm{\mathcal{G}}^i$ & $i$-th core (a third order tensor) of $\bm{\mathcal{G}}$ in TT-format\\
				\rowcolor{Light1} $\bm{\mathcal{G}}^i_{:,:,k}$ & $k$-th frontal slice (a matrix) of $\bm{\mathcal{G}}^i$ \\
				
				\rowcolor{Light2} $\Omega$ & Search space for ergodic exploration: $\Omega=[0,L]^d$ \\
				\rowcolor{Light2} $P(\bm{x})$ & Reference probability distribution defined on $\Omega$ \\
				\rowcolor{Light2} $\bm{\mathcal{P}}$ & Tensor formed by discretizing $P(\bm{x})$ \\
				\rowcolor{Light2} $\phi_k(x)$ & Elementary Fourier basis function for $[0,L]$, $\phi_k(x) = \cos({\frac{2 \pi (k-1) x}{L}})$  \\				
				\rowcolor{Light2} $\bm{\phi}(x)$ & A vector of elementary Fourier basis functions for $[0,L]$, $\bm{\phi}(x) = (\phi_1(x),\ldots,\phi_K(x))$  \\
				\rowcolor{Light2} $\bm{\Phi}(\bm{x})$ & The Fourier basis functions for $\Omega$ (a $d$-th order tensor) \\
				\rowcolor{Light2} $C_t(\bm{x})$ & Spatial statistics of the trajectory evolution $\bm{x}(t)$ of the dynamic system\\
				 \rowcolor{Light2} 
				 \rowcolor{Light2} 
				\rowcolor{Light2} $\bm{\mathcal{W}}_t$ & Fourier series coefficients of $C_t(\bm{x})$ \\

				\noalign{\hrule height 1.5pt}
			\end{tabular}
	}}
	\label{tab:notation}
\end{table}

\section{Problem Definition and Background}
\label{sec:problem}


In this section we lay out the mathematical background of our contribution. The notation used is summarized in Table \ref{tab:notation}.

\subsection{Tensors}
\label{tensors_define}
A tensor is a multidimensional array\footnote{The notion of tensors used in this paper is not to be confused with the one commonly used in physics as a multilinear map with specific properties. In this paper, a tensor is just a multidimensional array.}, whose order corresponds to the number of modes (or dimensions) of the array. A vector is a first order tensor and a matrix is a second order tensor. The $\bm{k}$-th element of a $d$-th order tensor $\bm{\mathcal{X}}\in\mathbb{R}^{K_{1}\times\cdots \times K_{d}}$, with indices $\bm{k}=(k_{1},\ldots,k_{d})$ and $k_{i} \in \{1,\ldots,K_i\}$, is denoted by $\bm{\mathcal{X}}_{\bm{k}}$. Here, $K_i \in \mathbb{Z}^{+}$ represents the size of $i$-th mode, with $i \in \{1,\ldots,d\}$. 

Fibers are the higher-order analogs of matrix rows and columns. A fiber is obtained by fixing every index but one. 

The inner product of two tensors $\bm{\mathcal{X}}$, $\bm{\mathcal{Y}} \in \mathbb{R}^{K_{1}\times\cdots\times K_{d}}$ is given by
\begin{equation*}
\langle\bm{\mathcal{X}}, \bm{\mathcal{Y}}  \rangle = \sum_{\bm{k}\in \mathcal{K}} \bm{\mathcal{X}_k} \bm{\mathcal{Y}_k},
\end{equation*}
where $\mathcal{K} = \{\bm{k}=(k_1,\ldots, k_d):  k_i \in \{1, \dots, K_i\}\}$.
The Frobenius norm of a tensor $\bm{\mathcal{X}}$ is defined by
\begin{equation*}
\| \bm{\mathcal{X}} \| = \langle \bm{\mathcal{X}},\bm{\mathcal{X}} \rangle^{\frac{1}{2}}.
\end{equation*}

The outer product of a tuple of one-dimensional arrays $( \bm{a}^1,\dots,\bm{a}^d )$, with $ \bm{a}^i=(a_1^i,\ldots,a_{K_i}^i)  \in \mathbb{R}^{K_i}$ is denoted by $ \bm{\mathcal{X}} = \bm{a}^1 \circ \cdots \circ \bm{a}^d \in \mathbb{R}^{K_1\times \cdots \times K_d}$, whose $\bm{k}$-th element is given by
\begin{equation*}
\bm{\mathcal{X}}_{\bm{k}} = a^1_{k_1} \cdots a^d_{k_d}.
\end{equation*}

In a wide variety of applications, observed data could be represented naturally as tensors \cite{tensors_survey_kolda,kroonenberg2008applied}. In this article tensors arise either due to the discretization of an underlying multivariate function or from the tensor representation of the variables involved in the ergodic control algorithm.



%
%
%
%
%

\subsection{Ergodic Control}
\label{erg_define}
Ergodic control considers a point-mass dynamical system whose trajectory evolves such that its time-averaged statistics matches a desired reference probability distribution. In the method proposed originally in \cite{mezic}, the problem reduces to minimizing a cost function called ergodic metric, evaluating the distance between the Fourier coefficients of the reference distribution and that of the time-averaged statistics of the trajectory evolution of the dynamical system. 



We assume a bounded $d$-dimensional rectangular domain: $\Omega = [0,L_{1}]\times\cdots\times[0,L_{d}]$ with $L_i > 0,  \forall i \in \{1,\ldots,d\}$. Without loss of generality, we will consider $L_i = L, \forall i \in \{1,..., d \}$.  $\bm{x}(t) \in \mathbb{R}^d$ represents the trajectory of the dynamical system in the domain. The spatial statistics of the trajectory $\bm{x}(t)$ is defined as the fraction of time spent by the dynamical system at each point of the domain:
\begin{equation*}
C_t(\bm{x}) = \frac{1}{t}\int_{\tau=0}^{t}\delta\big(\bm{x}(\tau)-\bm{x}\big) d\tau,
\end{equation*}
where $\delta$ is the Dirac delta function, and $\bm{x} = (x_{1},\ldots,x_{d}) \in \Omega$ is a point in the domain.

Let $P(\bm{x})$ be the reference probability distribution for the exploration defined on $\Omega$. The goal of ergodic control is to match the time-averaged spatial statistic $C_t(\bm{x})$ with the spatial distribution $P(\bm{x})$. The idea is to choose  $K\in \mathcal{Z}^{+}$ orthonormal Fourier basis functions\footnote{In general, we can choose a different number of basis functions $K_i$ for each dimension $i \in \{1,\dots,d \}$. Without loss of generality, we will assume $K_i = K, \forall i$.} satisfying the Neumann boundary conditions on the boundary of $\Omega$: $\phi_{k}$, $\forall k \in \{1,\ldots,K\}$, for each variable  $x$, which is then organized as $\bm{\phi}(x)=(\phi_{1}(x),\ldots,\phi_{K}(x))\in \mathbb{R}^{K}$. Although the results that follow apply to any such choice of basis function, we will use $\phi_k(x) = \cos({\frac{2 \pi (k-1) x}{L}})$ for numerical evaluation, see \cite{Calinon19MM} for details. Then, orthonormal Fourier basis functions for $\Omega$ can be obtained by the elements of the $d$-th order tensor formed by the outer product of these vectors: $\bm{\Phi}(\bm{x})=\bm{\phi}(x_{1})\circ\cdots\circ\bm{\phi}(x_{d}) \in \mathbb{R}^{K\times \cdots \times K}$. With respect to this basis, the Fourier coefficients (cosine transforms) of $P(\bm{x})$ can be represented by a $d$-th order tensor $\bm{\hat{\mathcal{W}}}$. For a given index $\bm{k}=(k_1,\ldots,k_d) \in \mathcal{K}$, we have $\bm{\Phi_k}(\bm{x}) = \phi_{k_1}(x_{1})\times\cdots\times\phi_{k_d}(x_{d})$, and $\bm{\hat{\mathcal{W}}}_{\bm{k}}$ represents the Fourier coefficient w.r.t.~the basis $\bm{\Phi}_{\bm{k}}$, namely
 \begin{equation}
 \label{eq:multi_int}
      \bm{\hat{\mathcal{W}}}_{\bm{k}} =  \int_{x_1=0}^L \cdots \int_{x_d=0}^L P(\bm{x})\bm{\Phi_k}(\bm{x}) dx_{1} \ldots dx_{d}.
 \end{equation}


The ergodic metric is then defined as 
\begin{equation}
    \label{eq:metric}
    \xi(t) = \sum_{\bm{k}\in \mathcal{K}}\bm{\Lambda_k}\big(\bm{\mathcal{W}}_{\bm{k}}(t)-\bm{\hat{\mathcal{W}}}_{\bm{k}} \big)^2,
\end{equation}
where $\bm{\Lambda_k} = (1+\|\bm{k}\|^2)^{-\frac{d+1}{2}}$ are the weights for different frequencies, $\mathcal{K} = \{\bm{k}=(k_1,\ldots, k_d):  k_i \in \{1, \dots, K\}\}$ and $K$ is a sufficiently large positive integer. This way, higher priority is given to lower frequency contents of the reference distribution (i.e., exploration of large scale features), hence resulting in a multi-scale exploration behavior. $\bm{\mathcal{W}}(t)$ is the Fourier coefficients for the spatial statistics of the trajectory evolution $\bm{x}(t)$ at time t (i.e., of $C_t(\bm{x})$), which is given by
\begin{equation}
    \bm{\mathcal{W}}(t)= \frac{1}{t}\int_{\tau=0}^t\bm{\Phi}(\bm{x}(\tau)) d\tau.
    \label{eq:w_t}
\end{equation}

The ergodic control objective is $ \lim_{t \to \infty} \xi(t) = 0$.
For a dynamical system $\dot{\bm{x}} = f(\bm{x},\bm{u})$ where $\bm{x} \in \mathbb{R}^{d}$, we want the ergodic dynamics w.r.t.~the evolution of the states $\bm{x}(t)\in \Omega \subset R^d$. For infinitesimal control horizon, the solution for first-order systems is given as (see \cite{mezic,Calinon19MM} for details)
\begin{align}
        \dot{x}_{i}(t) &= \alpha \frac{b_{i}(t)}{\|\bm{b}(t)\|},\nonumber\\
        \text{with}\quad b_{i}(t) &= \sum_{\bm{k}\in \mathcal{K}}\bm{\Lambda}_{\bm{k}}\big(\bm{\mathcal{W}}_{\bm{k}}(t)-\bm{\hat{\mathcal{W}}}_{\bm{k}}\big) \, \nabla_{i}\bm{\Phi}_{\bm{k}}\big(\bm{x}(t)\big),\nonumber\\
        \bm{b} &= (b_1,\ldots,b_d),\\
        \nabla_{i}\bm{\Phi}\big(\bm{x}(t)\big) &= \bm{\phi}(x_{1})\circ\cdots \circ \frac{\partial \bm{\phi}(x_{i})}{\partial x} \circ \cdots \circ\bm{\phi}(x_{d}),
        \label{eq:grad_phi}
\end{align}
where $\alpha>0$ is a small real number.

For a fully actuated system with point-mass dynamics $\dot{\bm{x}} = \bm{u}$, with $\bm{u}=(u_1,\ldots,u_d)$ and maximum velocity $u_{\max}$, the control commands $u_i, i \in \{1,\ldots,d\}$ that minimize the ergodic metric are given by
\begin{align*}
    u_{i}(t) &= u_{\max}\frac{b_{i}(t)}{\|\bm{b}(t)\|}.
\end{align*} 




Similar expressions for control policies are available for other types of dynamical systems, such as second-order point-mass systems (acceleration command), or first order and second order Dubin's car models, see e.g. \cite{mezic, mathew_experimental}. The controller can also be extended to other nonlinear dynamic systems, see e.g. \cite{ergodic_tutorial, miller_traj_opt, anastasia_erg}. We demonstrate our approach using the simple point-mass system, as it captures the key challenges in scaling the ergodic control algorithm to higher-dimensional exploration, and because it remains a classical choice for ergodic exploration \cite{mathew_chaotic_painting, mathew_experimental, mathew_uniform_coverage}.  

\begin{algorithm}
    \caption{Ergodic control algorithm}
    \label{alg:algo1}
    \begin{algorithmic}
        \STATE \textbf{Input:} $d$, $L$, $K$, $u_{\max}$, $T$, and $P(\bm{x})$
        \STATE \textbf{Preprocessing:}
        \STATE Compute  $\bm{\hat{\mathcal{W}}}$ (evaluate $K^d$ multivariate integrals with \eqref{eq:multi_int})
        \STATE Compute $\bm{\Lambda}$ ($K^d$ function evaluations)
      
        \STATE \textbf{Initialise:} $t=0$, $dt$ (time step), $\bm{x}(0)$, $\bm{\mathcal{W}}(0)$, $\bm{u}(0)$
       
        \STATE 
        \COMMENT{\emph{\color{gray}Control Loop}\footnotemark} 
        \WHILE{$t<T$}  
            \STATE $t \leftarrow t+dt$
            \STATE Update $\bm{x}(t)$ 
            \STATE Update $\bm{\mathcal{W}}(t)$ (use numerical integration)
            
            \STATE Compute $\nabla_i \bm{\Phi}\big(\bm{x}(t)\big), \forall i\in\{1,\ldots,d\}$ 
    
            \FOR{i=1,\ldots,d} 
            \STATE $b_{i}(t) = \sum_{\bm{k}\in \mathcal{K}}\bm{\Lambda}_{\bm{k}}\big(\bm{\mathcal{W}}_{\bm{k}}(t)-\bm{\hat{\mathcal{W}}}_{\bm{k}}\big) \, \nabla_{i}\bm{\Phi}_{\bm{k}}\big(\bm{x}(t)\big)$
            \ENDFOR
            \STATE Update $\bm{u}(t) = u_{\max} \frac{\bm{b}(t)}{\|\bm{b}(t)\|}$
        \ENDWHILE 
    \end{algorithmic}
\end{algorithm}
\footnotetext{In the control loop, each of the variables $\bm{\hat{\mathcal{W}}},\bm{\mathcal{W}}(t),\bm{\Lambda},\nabla_i\bm{\Phi}\big(\bm{x}(t)\big)$ need $\mathcal{O}(K^d)$ floating-point elements and each binary operation involving them has computational complexity $\mathcal{O}(K^d)$.}

The ergodic control algorithm (see Algorithm 1) is simple but it suffers from the curse of dimensionality when the dimension $d$ of the exploration domain $\Omega$ increases, which typically limits the use of the algorithm to problems of 2 or 3 dimensions. This is a drawback since many applications, in practice, are of dimensions $d>3$. For example, the end-effector of a robot manipulator has 6 DOF (position and orientation). A naive implementation of the above algorithm for an exploration in the task space of a manipulator is then not feasible in practice.
\\
The main challenges in the algorithm are listed below:
\begin{enumerate}
    \setlength{\itemsep}{4pt}
    \item\emph{Computation and storage of the Fourier series coefficients of the reference distribution $\bm{\hat{\mathcal{W}}}$:}\\[2pt]
        This requires evaluation of the multidimensional integral in \eqref{eq:multi_int} $K^d$ times to completely determine $\bm{\hat{\mathcal{W}}}$ . Although this is a preprocessing step, the computational complexity involved and the storage requirement make it infeasible for engineering applications.
    \item \emph{Computation and storage of $\bm{\Lambda}$:}\\[2pt]
        This is a preprocessing step. The computation of each element $\bm{\Lambda_k}$ is straightforward, but the number of function evaluations to find the complete tensor $\bm{{\Lambda}}$ and the required storage grow exponentially (i.e., $K^d$).
    \item \emph{Real-time implementation of the control loop:}\\[2pt]
        The control loop will be very slow as the algebraic operations (such as addition, element-wise product, summation, Frobenius norm of tensors) are more time-consuming as the order of the tensors involved in computing the control policy increases. So, a realtime implementation of the control loop may not be possible. 
\end{enumerate}

To give an example of the computation time involved in ergodic control, we used the Python software implementation of ergodic control described in \cite{mezic} \footnote{The configuration of computing system used is mentioned in Section \ref{numerical_evaluation}.}. By using $K=10$ for the Fourier series coefficients and a spherical Gaussian of variance $0.01$ at the center of $\Omega$ with $L=1$ as the reference probability distribution, the preprocessing time to compute the coefficients\footnote{Here, each coefficient is computed independently.} $\bm{\hat{\mathcal{W}}}$ is approximately $16$s in 2D, $5400$s in 3D, and $2300000$s in 4D ($d=4$) with the multivariate integration \eqref{eq:multi_int}  evaluated using the Python package \textit{scipy.integrate.nquad}.\footnote{ \url{https://docs.scipy.org/doc/scipy/reference/tutorial/integrate.html}} Note that the number of elements to be stored in these cases is $10^d$, and for larger $d$ (such as $d>6$) it is highly likely that memory/storage requirement for each of the multidimensional arrays involved exceeds the limit. Moreover, the average time taken per control loop in the above setting for $d=3,4,5,6,7$ are $7\times 10^{-4}$s, $1\times10^{-3}$s, $4\times10^{-3}$s, $6\times10^{-2}$s, and $8\times10^{-1}$s correspondingly. For $d=2$, \cite{dressel_info_maps} uses a supervised learning approach based on convolutional neural network for the fast computation of the Fourier coefficients. For higher dimensional problems,  as we will see next, the above-mentioned challenges can be solved using the capability of a state-of-the-art tensor decomposition technique called tensor train (TT). Using the proposed solution, for the above reference distribution with $d=7$, $\bm{\hat{\mathcal{W}}}$ can be represented using approximately $Kd$ parameters and it can be computed in less than $2\times10^{-3}$s and the average control loop takes less than $1\times10^{-3}$s. Also, for other reference distributions and larger $d$, the pre-processing step and the control loop can be processed very fast.

\subsection{Tensor Decomposition Techniques}
\label{tensor_decomposition_define}
Tensor decomposition (or tensor factorization) extends matrix decomposition techniques (e.g. matrix singular value decomposition) to multidimensional arrays. They allow tensors to be encoded compactly using a set of lower-order and/or lower-rank tensors (called factors). These factor elements are operated, depending on the decomposition technique, using various algebraic operations to represent elements of the given tensor. The accuracy of the representation is often controlled by the \textit{rank} of the tensor decomposition. The rank of a tensor has different meanings for different tensor decomposition techniques. If the tensor has some low-rank structure (often due to separability arising from symmetry or smoothness or parallel proportionality), the number of elements required to represent the given tensor using the decomposition techniques will be far fewer than the original tensor.\footnote{For example, in matrix singular value decomposition, the low-rank corresponds to the case when many of the singular values are zero or negligible.} In addition to compactly representing tensors, tensor decomposition allows efficient application of algebraic operations in the compressed format.\footnote{This is analogous to algebraic operations on large matrices in their decomposition such as SVD, LU, etc. If we consider the right multiplication of a matrix $\bm{A}\in \mathbb{R}^{m\times s}$ with $\bm{B}\in \mathbb{R}^{s \times n}$, and if we have a decomposition of $\bm{A}=\bm{Q}\bm{R}$ with $\bm{R}\in \mathbb{R}^{p \times n}$ and $p{\ll}s$, then computing $\bm{C} = \bm{Q}\bm{R}\bm{B}$ using the decomposition of $\bm{A}$ is more efficient than directly computing $\bm{C} = \bm{A}\bm{B}$ using the raw $\bm{A}$. Similarly, for scalar multiplication of $\bm{A}$, computing $\bm{C} = (k\bm{Q}) \bm{R}$ is more efficient than $\bm{C} = k \bm{A}$.}

Popular tensor decomposition techniques include CANDECOMP/PARAFAC (CP), also known as canonical polyadic, Tucker, hierarchical Tucker (HT), and tensor train (TT), see \cite{survey_kressner} for a survey. For $d=2$, all these techniques reduce to the well known singular value decomposition (SVD) of matrices. They can be interpreted as higher-order extensions of SVD, where these decomposition techniques differ for tensors of order $d > 2$. There exist many powerful algorithms to find the decomposition for each technique, most of them are based on the idea of alternating least squares (ALS). The CP decomposition of a tensor is formed by sum of rank-1 tensors. Although the number of parameters used by CP decomposition is small and scales well to higher order tensors, the space of  CP tensors with a finite maximal rank is not closed. This may result in an ill-posed problem to find the best low-rank approximation. The Tucker decomposition overcomes this issue but it suffers from the curse of dimensionality, see \cite{tensors_survey_kolda} for details.

The TT decomposition \cite{oseledets_tt} shares the good properties of the Tucker decomposition because the space of the tensors in TT format (with a fixed maximal rank) forms a smooth manifold, thus allowing robust algorithms to determine the TT decomposition of a tensor. It also shares good properties with CP, as it scales well to high-order tensors. We exploit the tensor train (TT) decomposition in our approach, as it gathers the good properties of both CP and Tucker decomposition. In the next section, a brief overview of TT decomposition is given.


\subsection{Tensor Train Decomposition}
\label{tt_define}
A $d$-th order tensor $\bm{\mathcal{G}} \in \mathbb{R}^{K_1\times\cdots \times K_d}$ in TT format is represented using a tuple of $d$ third-order tensors
$(\bm{\mathcal{G}}^1,\ldots,\bm{\mathcal{G}}^d)$.

\begin{figure}[t!]
    \centering
     \includegraphics[width=0.95\linewidth]{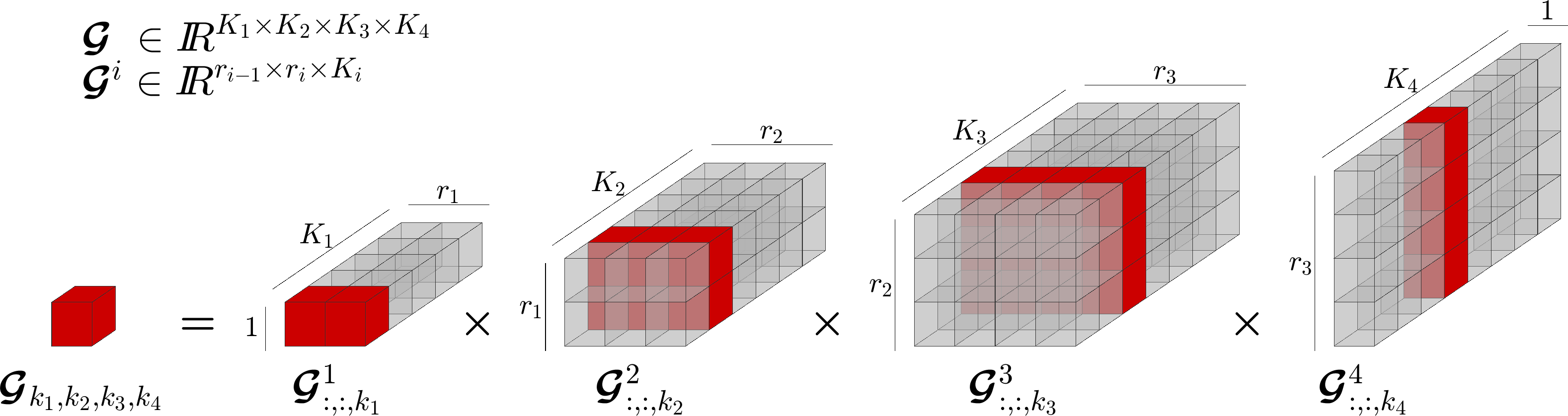}
     \caption{An element of a tensor in TT-format can be accessed by multiplying the selected slices (matrices represented in red color) of the core tensors (factors). The figure depicts an example for a 4th order tensor $\bm{\mathcal{G}}\in\mathbb{R}^{5\times6\times7\times8}$ of  rank $\bm{r}=(2,3,4)$.}
     \label{fig:tt_format}
\end{figure}

Here, $\bm{\mathcal{G}}^i\in \mathbb{R}^{r_{i-1} \times r_{i} \times K_i }$,  $i \in \{2,\ldots,d{-}1\}$, $\bm{\mathcal{G}}^1\in \mathbb{R}^{1 \times r_{1} \times K_1 }$ and
$\bm{\mathcal{G}}^d\in \mathbb{R}^{ r_{d-1} \times 1 \times K_{d}}.$ As shown in Fig. \ref{fig:tt_format}, the $\bm{k}$-th element, with  $\bm{k} \in \mathcal{K} =\{(k_{1},\ldots,k_{d})\colon k_i \in \{1,\ldots,K_i\}, i \in \{1,\ldots,d\} \}$, is given by  
\begin{equation}
    \label{eq:tt_rep}
    \bm{\mathcal{G}_k} = \bm{\mathcal{G}}^1_{:,:,k_1}\bm{\mathcal{G}}^2_{:,:,k_2}\cdot\cdot\cdot \bm{\mathcal{G}}^d_{:,:,k_d},
\end{equation}
where $\bm{\mathcal{G}}^i_{:,:,k_i} \in \mathbb{R}^{r_{i-1} \times r_i}$ represents the $k_{i}$-th frontal slice (a matrix) of the third order tensor $\bm{\mathcal{G}}^i$.
The \textit{TT-rank} of the tensor in TT representation is then defined as the tuple $\bm{r}=(r_{1},r_{2},\ldots,r_{d-1})$. We call $r$ = $\max{(r_1,\ldots,r_{d-1})}$ as the \textit{maximal TT rank}.  For any given tensor, there always exists a TT decomposition \eqref{eq:tt_rep}.

The TT decomposition of a tensor can be considered as a particular way of writing the elements of a given tensor as a finite sum of separable products \cite{survey_kressner}. To explain this intuitively, consider a multivariate function $g \colon \Omega_g \subset  \mathbb{R}^d \to \mathbb{R}$ such that $g(\bm{x})=g(x_1,\ldots, x_d)$ which can be approximated as a finite sum of products of one dimensional functions ($g_{\hat{r}_i}^{i} \colon \mathbb{R} \to \mathbb{R}, \text{ } i \in \{1,\ldots,d\} \text{ and } \hat{r}_i \in \{1,\ldots,r_i\}, r_i \in \mathbb{Z}^{+} $), namely
\begin{equation}
    \label{eq:separability}
    g(x_1,\ldots,x_d) \approx \sum_{\hat{r}_1=1}^{r_1}\cdots\sum_{\hat{r}_d=1}^{r_d}g_{\hat{r}_1}^1(x_1)\cdots g_{\hat{r}_d}^d(x_d).
\end{equation}

Suppose we have a discretization of the domain
$\Omega_g \subset \mathbb{R}^d$ 
given by the set 
$\{ \bm{y}=(y_{k_1},\ldots,y_{k_d}): y_{k_i}\in \mathbb{R}, k_i \in \{1,\ldots, K_i\}, K_i \in \mathbb{Z}^{+}\}$ 
with the corresponding index set
$\mathcal{K}$ as define above.
Let $\bm{\mathcal{G}}$ be the tensor formed by evaluating the function $g$ at these discretization points (i.e., $\bm{\mathcal{G}}_{\bm{k}} = g(y_{k_1},\ldots,y_{k_d}), \forall \bm{k}=(k_1,\ldots,k_d) \in \mathcal{K}$). Then, the TT decomposition of $\bm{\mathcal{G}}$ is analogous to the representation \eqref{eq:separability} of the underlying multivariate function $g$, and it is given by
\begin{equation*}
    \bm{\mathcal{G}}_{\bm{k}} = \sum_{\hat{r}_1=1}^{r_1}\cdots\sum_{\hat{r}_d=1}^{r_d}\bm{\mathcal{G}}_{1,\hat{r}_1,k_1}^1\cdots \bm{\mathcal{G}}_{\hat{r}_{d-1},1,k_d}^d, \quad \forall \bm{k} \in \mathcal{K}, 
\end{equation*}
which is compactly written in \eqref{eq:tt_rep} using matrix multiplication.

Let $r$  be the maximal TT-rank and $K = \max(K_{1},\ldots,K_{d})$. Then, the number of elements in the TT representation is $\mathcal{O}(Kdr^{2})$ as compared to  $\mathcal{O}(K^{d})$ elements in the original tensor. For small $r$, the representation is thus very efficient. In addition to storage efficiency, many numerical algebraic operations on tensors such as addition, scalar multiplication, Hadamard (or element wise) product,  Frobenius norm can be directly and efficiently applied in TT representation\footnote{For example, to gain an intuition about the efficiency of algebraic operations in TT format, consider two $d$-th order tensors of same shape $\bm{\mathcal{G}}$ and $\bm{\mathcal{H}}$  with known TT representations. The TT cores of the tensor $\bm{\mathcal{S}} = \bm{\mathcal{G}} + \bm{\mathcal{H}}$ can be directly obtained as  $\bm{\mathcal{S}}_{:,:,k}^i = \text{diag}(\bm{\mathcal{G}}_{:,:,k}^i,\bm{\mathcal{H}}_{:,:,k}^i)$  for $i \in \{2,\ldots, (d-1)\}$ and $\bm{\mathcal{S}}_{:,:,k}^j = \text{concatenate}(\bm{\mathcal{G}}_{:,:,k}^j,\bm{\mathcal{H}}_{:,:,k}^j)$ for $j \in \{1,d\}$, where \textit{diag} is the block diagonal matrix construction and \textit{concatenate} represents vector concatenation operation. Also, the TT cores of the tensor formed by scalar multiplication  $\bm{\mathcal{Q}} = c \cdot \bm{\mathcal{G}}$ can be obtained by multiplying the scalar only to the first core: $\bm{\mathcal{Q}}^1=c \cdot\bm{\mathcal{G}}^1$ and $\bm{\mathcal{Q}}^i= \bm{\mathcal{G}}^i$ for $i \in \{2, \ldots,d \}$.}. These operations are explained in \cite{oseledets_tt, CichockiTO}. Most algebraic operations on tensors in TT format have computational complexity linear in $d$ and $K$, and polynomial (often quadratic or cubic) in $r$. As we will see in Section \ref{main_algo}, the proposed algorithm for ergodic control in this paper will exploit this efficiency in algebraic operation offered by TT format to speed up the computation in the control loop.

As explained above, the existence of low-rank structure (i.e., a low maximal TT-rank) of a given tensor is closely related to the separability of the underlying multivariate function. Although separability of functions is not a very well understood problem, it is known that smoothness\footnote{By smoothness, we mean the degree of variation of the function across its domain. For example, a probability density function in the form of Gaussian mixture model (GMM) is considered to become less smooth as the number of mixture components (i.e., multi-modality) increases or the variance of the component Gaussians decreases (i.e., sharper peaks). More formally, the degree of smoothness can be defined using the properties of higher-order derivatives.} and symmetry of functions often induces better\footnote{By \textit{better}, we mean fewer low-dimensional functions in the sum of products representation.} separability of the functions. This has been exploited to solve PDEs and problems involving high-dimensional integration \cite{survey_kressner}, and it is used in this paper to solve the challenges in ergodic control. 

\begin{figure}[t!]
    \centering
     \includegraphics[width=0.95\linewidth]{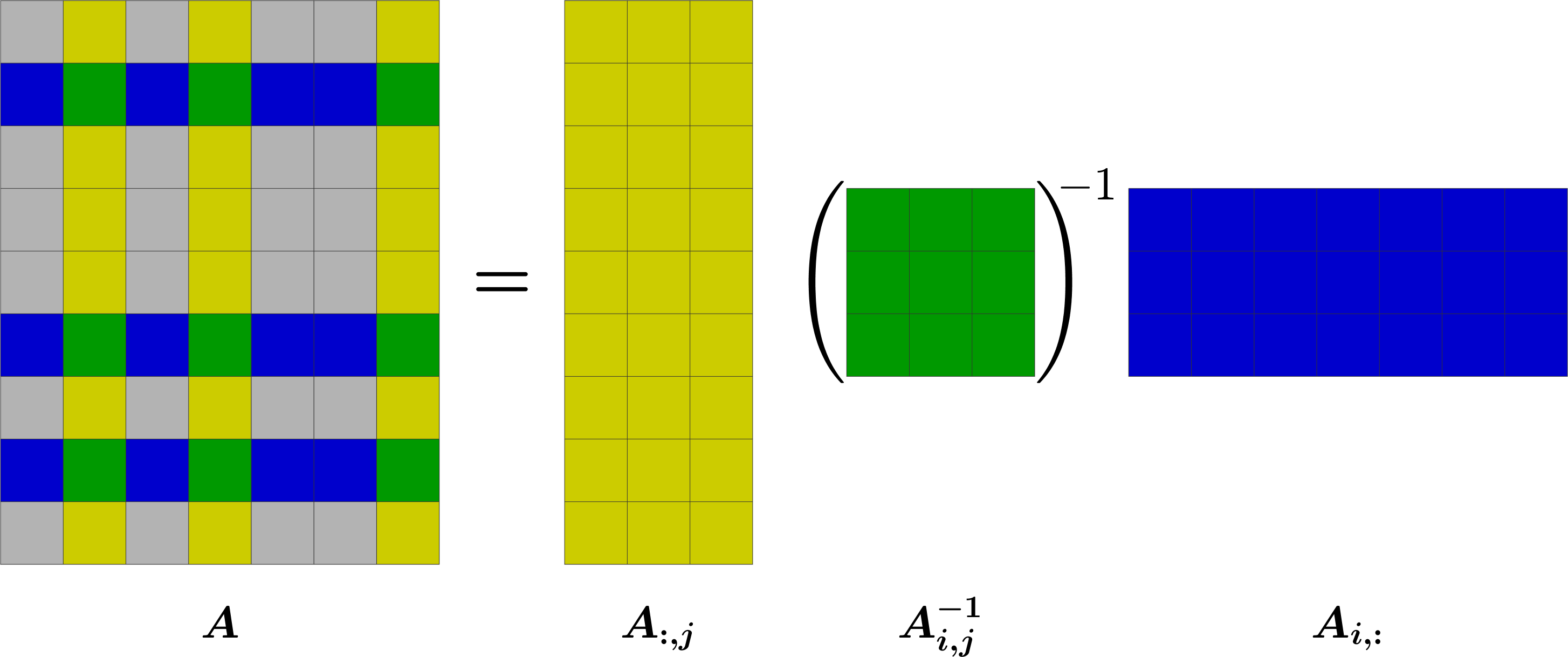}
     \caption{TT-cross is an extension of skeleton decomposition in matrix theory. A rank-$r$ matrix $\bm{A} \in \mathbb{R}^{m \times n}$ can be exactly recovered if we know $r$ independent columns of $\bm{A}$ indexed by $\bm{j} = (j_1,\ldots, j_r)$, $\bm{A}_{:,\bm{j}} \in \mathbb{R}^{m \times r}$ and $r$ independent rows of $\bm{A}$ indexed by $\bm{i} = (i_1,\ldots,i_r)$, $\bm{A}_{\bm{i},:} \in \mathbb{R}^{r \times n}$ of matrix $\bm{A}$, with their intersection $\bm{A}_{\bm{i},\bm{j}} \in \mathbb{R}^{r \times r}$ being nonsingular. Then, by skeleton decomposition we have $\bm{A} = \bm{A}_{:,\bm{j}} \bm{A}_{\bm{i},\bm{j}}^{-1} \bm{A}_{\bm{i},:}$.}
     \label{fig:cur}
\end{figure}

\newpage
\noindent\emph{Finding TT Decompositions}\\[2pt]
There exist many algorithms to find the TT decomposition of a tensor. The popular methods are TT-SVD, TT-DMRG, and TT-cross. TT-SVD and TT-DMRG, like matrix SVD, require the full tensor in memory to find the decomposition, and hence they are infeasible for higher order tensors. TT cross approximation (TT-cross) \cite{ttcross2}\cite{ttcross1} is an extension of the cross approximation technique (also called CUR or skeleton decomposition) in matrix theory \cite{survey_matrix_lowrank}, see Fig. \ref{fig:cur}. It is appealing for many practical problems as it approximates the given tensor with controlled accuracy, by using only a small number of its elements. When applied to tensors, the cross approximation method needs access to only certain fibers of the original tensor at a time, and hence works as a black box method.

Given a function (or a procedure) $g$ that evaluates an element of a tensor $\bm{\mathcal{G}}$ given its index and an approximation accuracy $\epsilon$ in the Frobenius norm, TT-cross returns an approximate tensor in TT format $\hat{\bm{\mathcal{G}}} = \text{TT-cross}(g,\epsilon)$ to the tensor $\bm{\mathcal{G}}$ by querying only a portion of its elements. The expected error in approximation is less than the accuracy $\epsilon$ specified.


Thus, TT-cross avoids the need to store explicitly the original tensor, which may not be possible for higher order tensors. It can approximate a $d$-th order tensor, with maximal TT-rank $r$ and size of each mode $K$, using only $\mathcal{O}(Kdr^2)$ samples and $\mathcal{O}(Kdr^3)$ operations (flops). This is very efficient if the TT-rank $r$ of the tensor is low, which is typically the case in many engineering applications, including robotics. In this paper, we will use TT-cross to find TT decompositions.
\\ 

\noindent\emph{TT-rounding}\\[2pt]
TT-rounding \cite{oseledets_tt} is an important operation on a tensor in TT format. Most binary operations on tensors in TT format, although  efficient, result in an increase in the TT-rank of the resultant tensor, where the resultant tensor is often not in its optimal TT representation. A repeated application of binary operations to a given tensor may result in an explosion of its TT-rank, which would effect the efficiency of subsequent operations on the tensor. For example, the addition of two tensors,
both with TT-rank $\bm{r} = (r_1,\dots, r_d)$, results in a tensor in TT format with rank $\bm{r} = (2r_1,\dots, 2r_d)$. TT-rounding is an operation applied to tensors already in TT format to compress it to optimal TT representation and hence reduces its TT-rank. For a $d$-th order tensor $\bm{\mathcal{G}}$ in TT-format with maximal TT-rank $r$, TT-rounding has computational complexity $\mathcal{O}(Kdr^3)$. The TT-rounding procedure returns a tensor $\hat{\bm{\mathcal{G}}} =\text{TT-round}(\bm{\mathcal{G}},\hat{r})$, for a given $\hat{r}<r$, such that its maximal TT-rank is less than $\hat{r}$ and the Frobenius norm of the residual $\|\hat{\bm{\mathcal{G}}} - \bm{\mathcal{G}}\|$ is as small as possible. Alternatively, we can specify an approximation accuracy $\epsilon$ to a tensor $\bm{\mathcal{G}}$ in TT format and the TT-rounding returns a tensor $\hat{\bm{\mathcal{G}}} = \text{TT-round}(\bm{\mathcal{G}},\epsilon)$ with optimal TT-rank and $\|\hat{\bm{\mathcal{G}}} - \bm{\mathcal{G}}\| \leq \epsilon$.

\section{Algorithm Description}
\label{main_algo}
In this section, we give details of the solution proposed in this paper to overcome the challenges mentioned in Section \ref{erg_define}. Additionally, as part of our proposed strategy for 6D exploration in manipulation tasks, we propose a Riemannian manifold extension to allow ergodic exploration for orientation data represented as unit quaternions.

We use the TT representation for the variables involved in the algorithm, namely $\bm{\hat{\mathcal{W}}}, \bm{\mathcal{W}}(t), \bm{\Lambda}, \nabla_i \bm{\Phi}\big(\bm{x}(t)\big)$.\footnote{By its definition in \eqref{eq:grad_phi}, $\nabla_{i} \bm{\Phi}\big(\bm{x}(t)\big), \forall i \in \{1,\ldots,d \}$ can be represented as a rank-1 TT. Its TT cores are $\big(\bm{\phi}(x_{1}),\ldots,\bm{\phi}(x_{i-1}), \frac{\partial \bm{\phi}(x_{i})}{\partial x},\bm{\phi}(x_{i+1}),\ldots, \bm{\phi}(x_{d})\big)$. Similarly, $\bm{\Phi}(\bm{x})$ by its definition can be represented as a rank-1 TT, with TT cores $\big(\bm{\phi}(x_{1}),\ldots,\bm{\phi}(x_{d})\big)$} These variables, as described below, can be compactly represented in TT format, within a desired accuracy. Since all the operations to find the control policy, at each step of the control loop, are done on variables in TT format, the computational complexity is significantly reduced. 

$\bm{\Lambda}$ can be found using the TT-cross approximation. Because of the involved symmetry (see the definition of $\bm{\Lambda}$), the resultant tensor in TT format is of very low rank. Maximal TT rank of 2 accurately captures the tensor with error less than $10^{-2}$ in Frobenius norm and it can be computed in a fraction of a second (for $d{<}15$). We also represent $\bm{\mathcal{W}}(t)$ in TT format and use time integration in TT format \cite{lubichttintegrate} to compute it efficiently at each iteration of the control loop. Due to integration, as the TT-rank of this tensor may increase in an unbounded manner over time, we specify an upper bound (a hyperparameter) on its TT-rank. If other integration schemes are used, one can periodically use TT-rounding to cut off the TT-rank.

Finding $\bm{\hat{\mathcal{W}}}$ is a preprocessing step for the algorithm, and it is the most challenging part as it requires to evaluate $K^d$ times a $d$-dimensional integral given by \eqref{eq:multi_int}, if implemented naively. By using the properties of the TT format, the Fourier Coefficients can be computed in the following ways:\\[2pt]
    \emph{Method 1: By exploiting the properties of Gaussian mixture model (GMM) and the efficiency of TT-cross for reference distributions in the form of a GMM.}\\[2pt]
    By using TT-cross the above integral needs to be computed only at the query points (Fourier coefficients) of the TT-cross, thus we obtain a significant savings in the number of evaluations of the multidimensional integral. This will require the integral in \eqref{eq:multi_int} to be evaluated $\mathcal{O}(Kdr^2)$ times and $r$ is often small due to the structure of $\bm{\hat{\mathcal{W}}}$. However, computation of each multidimensional integral is still time consuming.
    For reference distributions in the form of a GMM, we have presented an analytical expression for the above integral in \cite{Calinon19MM}. This avoids computation of the multi-dimensional integration using numerical schemes for distributions in the form of GMM for reasonably\footnote{Here, we consider $d$ approximately up to 10 as the computation complexity of each Fourier coefficient using the analytical solution provided in \cite{Calinon19MM} grows in proportion to $2^d$. For details, refer to  \cite{Calinon19MM}.} large $d$ as we use the analytical expression to compute the Fourier coefficients at the query points of TT-cross. Thus the method directly exploits the structure (i.e., smoothness) of the Fourier coefficients in $\bm\hat{{\mathcal{W}}}$.\\[2pt]
    \emph{Method 2: By exploiting the properties of TT decomposition for arbitrary reference distributions.}\\[2pt] 
    We provide a solution to the Fourier coefficients $\bm{\hat{\mathcal{W}}}$ without having to perform any multidimensional integration directly. By exploiting the properties of Gaussian quadrature rule for the integration of multivariate functions, we derive an analytical expression for the Fourier coefficients $\bm{\hat{\mathcal{W}}}$ in the TT representation for arbitrary functions. In this method, we exploit the smoothness of the reference probability distribution $P(\bm{x})$ to find $\bm{\hat{\mathcal{W}}}$ indirectly by evaluating $P(\bm{x})$ at a few points in its domain. This will be the method we use in the rest of the paper for finding $\bm{\hat{\mathcal{W}}}$ unless otherwise stated. This method to find Fourier coefficients is very efficient for smooth high-dimensional functions, which will be explained in the next section. 

Note that, in practice,  the reference distributions are smooth and/or the Fourier coefficients vary smoothly across their indices. Hence, a low-rank TT representation of $\bm{\hat{\mathcal{W}}}$ is expected.  For GMM with very sharp peaks, one may prefer Method 1 as it relies directly on the structure of $\bm{\hat{\mathcal{W}}}$ and hence it could be faster for such non-smooth reference distributions in the form of GMM. However, the computationally complexity involved in evaluating the analytical expression for each Fourier series coefficient in Method 1 grows in proportion to $2^d$ and linearly with the number of mixture components in the GMM \cite{Calinon19MM}. Moreover, Method 1 requires the GMM to have negligible tail outside the domain of exploration $\Omega$. Since Method 2 applies to arbitrary reference distributions and it is observed in practice to be equally fast compared to Method 1 for smooth GMM as reference distribution, in the remaining sections we consider only this method to find the Fourier series coefficients $\bm{\hat{\mathcal{W}}}$.

\subsection{Finding the Fourier Series Coefficients}
\label{main_fourier}
In this section, we provide an analytical expression for the Fourier coefficients of a smooth but arbitrary distribution introduced in method 2 of Section \ref{main_algo}. The proof is inspired by \cite{stt} where they have used separability structure in TT representation to find polynomial approximations of multivariate functions. A similar strategy has been used in  \cite{dolgov_integration} to evaluate high-dimensional integration of smooth functions. We use this strategy to find an analytical expression for the Fourier coefficients of functions directly in TT representation. The proof relies on quadrature rules (see Appendix \ref{appdx_1} for the details) for numerical integration of multivariate functions. There are many possible options to choose the quadrature rule depending on the type of function $P(\bm{x})$ to be integrated. The following result applies to any choice of quadrature rule, however, for simplicity, we use Gaussian quadrature rule (G-Q) in this paper. 

The idea is to find the TT representation $\bm{\mathcal{P}}$ of the multivariate function $P(\bm{x})$ evaluated at the discretization induced from the quadrature rule. Then, as we will see below, the TT representation of $\bm{\hat{\mathcal{W}}}$ can be obtained directly  using $\bm{\mathcal{P}}$.
    
Let $x_j \in \mathbb{R}$ be the discretization points of the interval $[0,L]$ and $\alpha_j$ be the weights obtained from the quadrature rule, where $j \in \{ 1,\ldots, N\}$, with $N$ representing the specified degree of approximation of the function. Then, we can discretize the domain $\Omega$ at $\bm{x_j} = (x_{j_1},\ldots, x_{j_d})$, with $\bm{j} \in \mathcal{J}$, where $\mathcal{J} = \{\bm{j}=(j_1,\ldots, j_d):  j_i \in \{1, \dots, N\}\}$ is the index set. Let $\bm{\mathcal{P}}$ be the tensor formed by evaluating the reference distribution $P(\bm{x})$ at the discretization points, so that  $\bm{\mathcal{P}_j} = P(\bm{x_j}),  \forall \bm{j} \in \mathcal{J}$. 

Let $(\bm{\mathcal{P}}^1, \bm{\mathcal{P}}^2, \ldots, \bm{\mathcal{P}}^d)$ be the TT cores of $\bm{\mathcal{P}}$ in its TT representation\footnote{This can be obtained using TT-cross: $\bm{\mathcal{P}} = \text{TT-Cross}(P(\bm{x}), \epsilon)$.}, so that for $\bm{j}=(j_1,\ldots,j_d) \in \mathcal{J}$ we have,
\begin{equation*}
\bm{\mathcal{P}}_{\bm{j}} = \bm{\mathcal{P}}^1_{:,:,j_1} \bm{\mathcal{P}}^2_{:,:,j_2}\cdots \bm{\mathcal{P}}^d_{:,:,j_d},
\end{equation*}
then the TT cores of $\bm{\hat{\mathcal{W}}}$ are (see Appendix \ref{appdx_1} for the proof)
\begin{equation}
    \label{W_expression}
    \bm{\hat{\mathcal{W}}}^i_{:,:,k}\! = \sum_{j=1}^N \alpha_{j} \, \bm{\mathcal{P}}^i_{:,:,j} \, \phi_{k}(x_{j}), 
    \begin{array}{l}
    \forall k \in \{1,\ldots,K\},\\ 
    \forall i \in \{1,\ldots,d\},
    \end{array}
\end{equation}
so that 
\begin{equation*}
     \bm{\hat{\mathcal{W}}}_{\bm{k}} = \bm{\hat{\mathcal{W}}}^1_{:,:,k_1} \cdots \bm{\hat{\mathcal{W}}}^d_{:,:,k_d}, \; \forall \bm{k} \in \mathcal{K}.
\end{equation*}    

Thus, we can compute the Fourier coefficients $\bm{\hat{\mathcal{W}}}$ by only investing in computing the TT decomposition $\bm{\mathcal{P}}$ of the discretized reference distribution. This can be done in $\mathcal{O}(Ndr^2)$ function evaluations of $P(\bm{x})$ using the TT-cross algorithm. The TT-rank of the tensor $\bm{\hat{\mathcal{W}}}$ will be same as that of the TT-rank of $\bm{\mathcal{P}}$. For a smooth reference distribution, $\bm{\mathcal{P}}$ will have low TT-rank. This is a tremendous saving in computing the Fourier coefficients $\bm{\hat{\mathcal{W}}}$, and thus it overcomes the curse of dimensionality. The TT based algorithm for ergodic control is outlined in Algorithm 2.\footnote{Our implementation of Algorithm 2 in Python can be found at \url{https://sites.google.com/view/ergodic-exploration/}.}

\begin{algorithm}
\caption{Ergodic Control using TT}
\label{algo2}
\begin{algorithmic}
\STATE \textbf{Input:} $d$, $L$, $K$, $u_{\max}$, $T$, and $P(\bm{x})$
\STATE \textbf{Pre-Processing:}
\STATE Compute $\bm{\mathcal{P}}$ using TT-cross  
\STATE Find $\bm{\hat{\mathcal{W}}}$ using \eqref{W_expression}
\STATE Compute $\bm{\Lambda}$ using TT-cross
\STATE TT-rounding of $\bm{\hat{\mathcal{W}}}$ (remove low-energy contents)
\STATE \textbf{Initialise:} $dt$ (time step), $\bm{x}(0)$, $\bm{\mathcal{W}}(0)$ in TT format, $\bm{u}(0)$, and $t=0$
\STATE 
\COMMENT{\emph{\color{gray}Control Loop}\footnotemark}   
\WHILE{$t<T$}  
    \STATE $t \leftarrow t+dt$
    \STATE Update $\bm{x}(t)$ (time integration)
    \STATE Update $\bm{\mathcal{W}}(t)$ (Use TT time integration \cite{lubichttintegrate} with a fixed maximal TT-rank. Alternatively, use numerical integration such as Euler integration followed by TT-rounding)
    \STATE Compute $\nabla_{i} \bm{\Phi}\big(\bm{x}(t)\big), i \in \{1,\ldots,d \}$(rank-1 TT)
    \STATE \COMMENT{\emph{\color{gray}Using algebraic operations in TT}}
    \FOR{i=1,\ldots,d} 
    \STATE $b_{i}(t) = \sum_{\bm{k}\in \mathcal{K}}\bm{\Lambda}_{\bm{k}}\big(\bm{\mathcal{W}}_{\bm{k}}(t)-\bm{\hat{\mathcal{W}}}_{\bm{k}}\big)  \nabla_{i}\bm{\Phi}_{\bm{k}}\big(\bm{x}(t)\big)$
    \ENDFOR
    \STATE Update $\bm{u}(t) = u_{\max} \frac{\bm{b}(t)}{\|\bm{b}(t)\|}$
\ENDWHILE 
\end{algorithmic}
\end{algorithm}

\subsection{Ergodic Control on Riemannian Manifolds}
\label{sec:mani}
Most manipulation tasks concern the full robot end-effector pose, which includes both its position and orientation. Hence, when designing exploration strategies for manipulation it is desirable to consider both. In the case of position, the ergodic control formulation in Section \ref{main_fourier} can be directly applied. However, since orientation data do not lie on a Euclidean space, exploration in orientation requires a special mathematical treatment. In this section, we extend ergodic control to handle data on a Riemannian manifold \cite{Zeestraten17, Calinon20RAM}, particularly the orientation manifold $\mathcal{S}^3$.

\footnotetext{In the control loop, the memory of each variable and the computational complexity of each algebraic operation has complexity grow linearly with $d$. Thus it avoids the curse of dimensionality. In particular, the computation of $b_i(t)$ requires a subtraction, a Hadamard product and an inner product involving tensors in TT format, hence it can be computed efficiently.}

The orientation of the robot end-effector can be represented by a unit quaternion $\bm{q} \in \mathcal{S}^3$, comprised of a scalar part $q_s\in\mathbb{R}$ and a vector part $\bm{q}_v\in\mathbb{R}^3$ such that $\bm{q}=\left[q_s\>\>\bm{q}^\trsp_v\right]^\trsp$. For any point on the manifold $\bm{g} \in \mathcal{S}^3$ there exists a tangent space $\mathcal{T}_{\bm{g}}\mathcal{S}^3$ in which standard Euclidean methods can be applied to orientation. The function that maps a quaternion $\bm{q}$ from the manifold to a tangent space is called the \textit{logarithmic map} and is given by
\begin{equation}
	\mathrm{Log}(\bm{q}) = \begin{cases} 
	\mathrm{acos}_*(q_s) \frac{\bm{q}_v}{\|\bm{q}_v\|}, & q_s \neq 1 \\
	 \left[0, 0, 0\right]^\trsp, &  q_s = 1 \end{cases},
	 \label{eq:logmap_origin}
\end{equation}
where $\mathrm{acos}_*(\cdot)$ is a modified arc-cosine function \cite{Zeestraten17}. Equation \eqref{eq:logmap_origin} maps $\bm{q}$ to the tangent space of the manifold origin. The mapping of $\bm{q}$ to the tangent space of an arbitrary point $\bm{g}$ is given by \cite{Zeestraten17}
\begin{equation}
	\mathrm{Log}_{\bm{g}}(\bm{q}) = 	\mathrm{Log}(\bar{\bm{g}}*\bm{q}),
	 \label{eq:logmap}
\end{equation}
where $\bar{(\cdot)}$ and  $*$ denote the quaternion conjugate and product, respectively. The logarithmic map represents a unit quaternion $\bm{q}$ as a 3-dimensional Euclidean vector $\bm{v}\in\mathbb{R}^3$. Quaternions can be retrieved from the  tangent space through the \textit{exponential map}
\begin{equation}
	\mathrm{Exp}(\bm{v}) = \begin{cases} 
	\left[\mathrm{cos}(\|\bm{v}\|), \; \mathrm{sin}(\|\bm{v}\|)\frac{\bm{v}^\trsp}{\|\bm{v}\|}\right]^\trsp, & \|\bm{v}\|\neq 0 \\
	 \left[1, 0, 0, 0\right]^\trsp, &  \|\bm{v}\| = 0 \end{cases},
	 \label{eq:expmap_origin}
\end{equation}
which, analogously to \eqref{eq:logmap}, maps from an arbitrary tangent space $\mathcal{T}_{\bm{g}}\mathcal{S}^3$ back to the manifold through 
\begin{equation}
	\mathrm{Exp}_{\bm{g}}(\bm{v}) = 	\bm{g}*\mathrm{Exp}(\bm{v}).
	 \label{eq:expmap}
\end{equation}

Given a set of unit quaternions (e.g. obtained from demonstrations), we  formulate orientation-ergodic control by modeling their distribution in the tangent space of their mean.
For $M$ end-effector orientations $\{\bm{q}_i\}^{M}_{i=1}$, the mean on the manifold $\bm{\mu}\in \mathcal{S}^3$ is computed iteratively with (see \cite{Zeestraten17, Calinon20RAM} for details)
\begin{equation}
    \bm{v} = \frac{1}{M}\sum^M_{i=1}\mathrm{Log}_{\bm{\mu}}(\bm{q}_i), \qquad \bm{\mu} \leftarrow \mathrm{Exp}_{\bm{\mu}}(\bm{v}).
    \label{eq:Riemannian_mean}
\end{equation}
All quaternions in the dataset can thus be mapped to the tangent space of the mean through $\{\bm{v}_i\}^M_{i=1} = \{\mathrm{Log}_{\bm{\mu}}(\bm{q}_i)\}^M_{i=1}$, allowing the proposed ergodic exploration (Algorithm 2) to be performed in $\mathbb{R}^3$, even for orientation. As desired orientations are computed, in the tangent space,  at each time step by $\hat{\bm{v}}(t) = \bm{v}(t)+\bm{u}(t)dt$, the exponential map generates a desired unit quaternion for the robot to track, using
\begin{equation}
	\hat{\bm{q}}(t) = \mathrm{Exp}_{\bm{\mu}}(\hat{\bm{v}}(t)).
	 \label{eq:expmap_tracking}
\end{equation}
In this way, ergodic control for end-effector poses is formulated as a 6D problem, where the first 3 dimensions correspond to position and the last 3 to orientation.

\section{Numerical Evaluation}
\label{numerical_evaluation}


\begin{table*}[!t]
\centering
\renewcommand{\arraystretch}{1.25}
\begin{tabular}{@{\extracolsep{2pt}}ccccccccccc}
	\toprule[0.12em]
    \multicolumn{3}{c}{} & \multicolumn{6}{c}{\textbf{Gaussian mixture model}} & \multicolumn{2}{c}{\multirow{2}{*}{\textbf{Uniform distribution}}}  \\ \cline{4-9}
    \multicolumn{3}{c}{}&\multicolumn{2}{c}{2 components}& \multicolumn{2}{c}{4 components} & \multicolumn{2}{c}{6 components} & \multicolumn{2}{c}{}    \\ 
    \cline{4-5} \cline{6-7}  \cline{8-9}  \cline{10-11}
    \multicolumn{3}{c}{} & With TT & Without TT & With TT & Without TT & With TT & Without TT & With TT & Without TT \\ \hline
   \multirow{5}{*}{\textbf{5D}}&\multirow{3}{*}{\# parameters}& $\nabla_{i} \bm{\Phi}$ & $50$&$10^5$ & $50$&$10^5$ & $50$&$10^5$ & $50$&$10^5$ \\
   & & $\bm{\Lambda}$ & $160$&$10^5$ & $160$&$10^5$ & $160$&$10^5$ & $160$&$10^5$ \\
   & & $\bm{\hat{\mathcal{W}}}$ & $160$&$10^5$ & $548$&$10^5$ & $1032$&$10^5$ & $50$&$10^5$ \\ 
   &\multicolumn{2}{c}{Average time per  loop} & $2\!\times\!\!10^{-3}$&$4\!\times\!\!10^{-3}$ & $3\!\!\times\!\!10^{-3}$&$4\!\times\!\!10^{-3}$ & $3.6\!\times\!\!10^{-3}$&$4\!\times\!\!10^{-3}$ & $2\!\times\!\!10^{-3}$&$4\!\times\!\!10^{-3}$ \\ 
   &\multicolumn{2}{c}{Pre-processing time} & $0.2$&$-$ & $0.87$&$-$ & $2.4$&$ -$ & $33\!\times\!\!10^{-3}$&$-$ \\ \hline
   
  \multirow{5}{*}{\textbf{6D}}&\multirow{3}{*}{\# parameters}& $\nabla_{i} \bm{\Phi}$ & $60$&$10^6$ & $60$&$10^6$ & $60$&$10^6$ & $60$&$10^6$ \\
   & & $\bm{\Lambda}$ & $200$&$10^6$ & $200$&$10^6$ & $200$&$10^6$ & $200$&$10^6$ \\
   & & $\bm{\hat{\mathcal{W}}}$ & $200$&$10^6$ & $695$&$10^6$ & $1431$&$10^6$ & $60$&$10^6$ \\ 
   &\multicolumn{2}{c}{Average time per  loop} & $3.2\!\times\!\!10^{-3}$&$63\!\times\!\!10^{-3}$ & $4.6\!\times\!\!10^{-3}$&$63\!\times\!\!10^{-3}$ & $5.4\!\times\!\!10^{-3}$&$63\!\times\!\!10^{-3}$ & $3\!\times\!\!10^{-3}$&$63\!\times\!\!10^{-3}$ \\ 
   &\multicolumn{2}{c}{Pre-processing time} & $30\!\times\!\!10^{-3}$&$-$ & $1.26$&$-$ & $3.6$&$-$ & $40\!\times\!\!10^{-3}$&$-$ \\ \hline
   
  \multirow{5}{*}{\textbf{7D}}&\multirow{3}{*}{\# parameters}& $\nabla_{i} \bm{\Phi}$ & $70$&$10^7$ & $70$&$10^7$ & $70$&$10^7$ & $70$&$10^7$ \\
   & & $\bm{\Lambda}$ & $240$&$10^7$ & $240$&$10^7$ & $240$&$10^7$ & $240$&$10^7$ \\
   & & $\bm{\hat{\mathcal{W}}}$ & $233$&$10^7$ & $860$&$10^7$ & $1801$&$10^7$ & $70$&$10^7$ \\ 
   &\multicolumn{2}{c}{Average time per  loop} & $4.8\!\times\!\!10^{-3}$&$0.8 $ & $6.8\!\times\!\!10^{-3}$&$ 0.8$ & $7.5\!\times\!\!10^{-3}$&$0.8$ & $3.6\!\times\!\!10^{-3}$&$0.8$ \\ 
   &\multicolumn{2}{c}{Pre-processing time} & $35\!\times\!\!10^{-3}$&$ -$ & $1.53$&$-$ & $4.9$&$-$ & $43\!\times\!\!10^{-3}$&$-$ \\
    \bottomrule[0.12em]
\end{tabular}
\caption{Computational speed and storage requirements in ergodic control for different reference probability distributions with $K=10$ and $L=1$m. The components of GMM are spherical Gaussians with $0.005$ variance. All the tensors in TT format are approximated with an accuracy of $10^{-2}$ in the Frobenius norm. The preprocessing time for the naive approach (without using TT) is not given in the table as it is computationally infeasible in the computing device used for the experiment.}
\label{tab:performance_tt}

\vspace{-0.5cm}
\end{table*}

%
%

%


In this section, we demonstrate the computational efficiency of the TT-based algorithm for ergodic control through simulations. We use Method 2 described in Section \ref{main_algo} to compute the Fourier series coefficients. The simulations are performed on a Lenovo Thinkpad personal computer with Intel(R) Core(TM) i7-8565U CPU at 1.80GHz with 24GB RAM. We use ttpy, a Python-based toolbox for working with TT.\footnote{\url{https://github.com/oseledets/ttpy}}

A naive implementation without using tensor decomposition techniques would require $K^d$ elements to store each of the tensors: $\bm{\hat{\mathcal{W}}}$, $\bm{\Lambda}$, and $\nabla_{i} \bm{\Phi}\big(\bm{x}\big)$ ($i \in \{1,\ldots,d \}$). However, a TT representation requires  less than $4Kd$ elements for $\bm{\Lambda}$ (with approximation error $10^{-2}$ in the Frobenius norm) and only $Kd$ elements to exactly represent $\nabla_{i} \bm{\Phi}\big(\bm{x}\big)$. As $\nabla_{i} \bm{\Phi}\big(\bm{x}\big)$ is a rank-$1$ tensor with explicit analytical expressions for its TT cores, it can be computed very fast. Computing the weights $\bm{\Lambda}$ can be done in a fraction of a second for $d \leq 15$. 

\begin{figure}[t!]
    \centering
     \includegraphics[width=1.0\columnwidth]{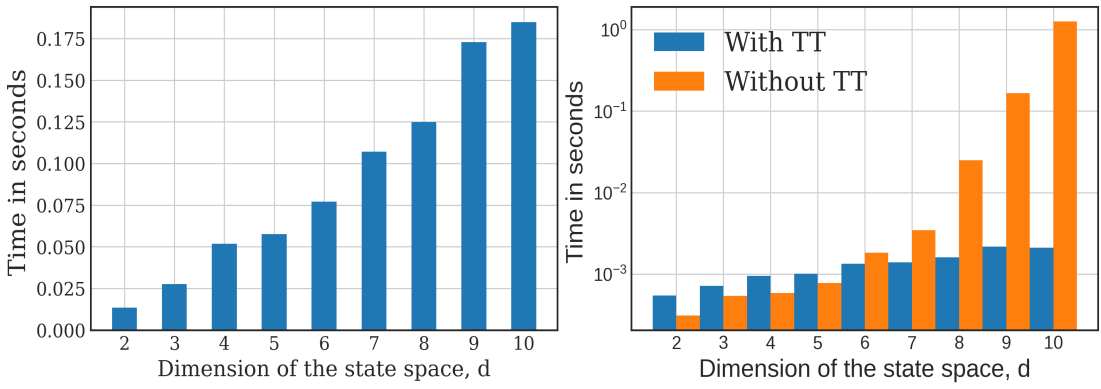}
     \caption{Time taken (linear scale) to compute the Fourier coefficients $\bm{\hat{\mathcal{W}}}$ (\textbf{left}) and the average time taken (in log-10 scale) per control loop (\textbf{right}) of the ergodic control algorithm using the proposed technique for $K=5$, $l=1$, and a reference distribution in the form of an isotropic Gaussian with variance $0.015$. In the right figure, the exponential growth in the computational complexity in the control loop can be observed for the standard approach (without using tensor decomposition), whereas the proposed approach avoids the curse of dimensionality. In the left figure, it can be observed that the computational complexity in computing the Fourier series coefficients using the proposed approach tends to grow linearly with $d$. }
     \label{fourier_nd_ctrl_loop}
\end{figure}

The computation and storage of $\bm{\hat{\mathcal{W}}}$ depends on the smoothness of the reference probability distribution. For the evaluation, we define our reference distribution as an isotropic Gaussian distribution at the centre of the domain with variance $0.015$, where we used $L=1$, $K=5$, $N=10$ and an approximation accuracy of $10^{-2}$ in the TT representation of $\bm{\hat{\mathcal{W}}}$ and $\bm{\Lambda}$. As can be seen in Fig. \ref{fourier_nd_ctrl_loop}, the time taken to compute $\bm{\hat{\mathcal{W}}}$ grows approximately linearly with $d$, and it is less than a second for the chosen reference distribution with $d \leq 10$ and the average time taken per control loop increases almost linearly with the number of dimensions $d$, whereas with a naive implementation (without using TT) the time taken in the control loop grows exponentially with $d$. The trend remains the same for other reference distributions, see Table \ref{tab:performance_tt}.

The computation of $\bm{\mathcal{W}}(t)$ in the control loop requires some attention. At each iteration of the control loop, the rank of the tensor $\bm{\mathcal{W}}(t)$ may keep increasing due to the integration, see \eqref{eq:w_t}. This could be a problem if the time period of ergodic exploration is very high. So, it is necessary to upper bound the TT-rank of this tensor using TT-rounding with a specified maximal TT-rank. Setting an excessively low value for the maximal rank may lead to convergence issues and specifying a large value for maximal rank slows down the speed of computation of each control loop. Thus, this hyperparameter must be chosen carefully. In the numerical evaluation, our experiments revealed that a maximal rank of $d\cdot r$ for $\bm{\mathcal{W}}(t)$, where $r$ is maximal TT-rank of $\bm{\hat{\mathcal{W}}}$, worked well for $d<15$.


The TT representation allows compact representation of the tensors involved and the storage complexity also grows linearly with $d$. These properties allow our algorithm to be implemented for real-time applications on devices with small memory and limited computational power, which are often the case in robotic systems. Another important property of our algorithm is that the approximation of the tensors involved such as $\bm{\hat{\mathcal{W}}}$, $\bm{\mathcal{W}}(t)$ and $\bm{\Lambda}$ can be controlled precisely using TT-rounding. In practice, TT-rounding with accuracy $10^{-2}$ is observed to be sufficient for all practical purposes considered here. Furthermore, doing these approximations in the spectral domain results in negligible impact on the time domain behaviour of the system, thanks to Parseval's theorem as the approximation can be considered as a noise filter in the ergodic motion. This also provides a convenient trade-off between accuracy of approximation in ergodic exploration and the speed of computation in the control loop.

\section{Experiment: Sensorless Peg-in-Hole Insertion using Ergodic Exploration}
\label{experimental_evaluation}
We evaluate the proposed approach in an insertion task. We formulate the insertion task as a 6D exploration problem where we simultaneously address the uncertainty about the insertion pose in position (3D) and orientation (3D) in the robot task space. Our method is well suited for peg-in-hole insertion tasks where uncertainties may arise from several sources, including variable grasps of the peg, unprecise locations of the hole, and unmodeled manufacturing defects of the involved components (gripper fingers, pegs and holes). In the considered experiment, the reference probability distribution for exploration is found using information from human demonstrations. The human demonstrates the key regions for exploration in the state space of the end-effector 
and we use a Gaussian mixture model (GMM) to model the reference probability distribution based on the datapoints collected during the demonstrations. As a means to intuitively show the effectiveness of our approach, we begin by comparing it to three baselines of exploratory behaviors  commonly used  in insertion tasks. For this we use a toy example in 2D and 3D.

\subsection{Simulation experiments}
\label{why_ergodic_insertion}


In this section we provide the motivation for using ergodic control for exploration, and its significance for insertion tasks, using simulation of exploration behavior in 2D and 3D. 
%

We use a GMM as the reference distribution in the space $\Omega$ with $L=1$m. The GMM is chosen such that it has $6$ equally weighted mixture components with its centers placed randomly in the exploration space $\Omega$ and each component is a spherical Gaussian with variance $0.01$. As a first metric, we compute the average time taken to reach a spherical region with volume $0.5\%$ of the volume of $\Omega$.  The spherical region is representative of the target detection region  during exploration. For the insertion task, this corresponds to the set of end-effector states at which the peg is inside the hole. For all the trials, we fix a maximum duration of $1000$s for detection (i.e., reaching the target region) and the magnitude of the point-mass-system velocity is constant and fixed to $u_{\max} = 0.1$ m/s. For the analysis we choose $10$ different GMMs as described above and for each choice of GMM, we conduct $10$ trials. For each trial, the center of the target region is chosen by sampling in $\Omega$ from the reference GMM and the point-mass system starts with the same initial state: $(0.5,0.5)$ for 2D and $(0.5,0.5,0)$ for 3D.

We compare four different exploration strategies: 
\begin{enumerate}
    \setlength{\itemsep}{3pt}
    \item\textbf{Strategy 1:} \emph{Ergodic exploration} (proposed approach),
    \item\textbf{Strategy 2:} \emph{Sampling-based exploration}
    \item\textbf{Strategy 3:} \emph{Cylindrical spiral}, as a representative of sweeping patterns,
    \item\textbf{Strategy 4:}
    \emph{Mixture of ellipsoidal spirals}, as a sweeping pattern customized for GMM.
\end{enumerate}

\begin{figure}
	\centering
	\begin{subfigure}{0.75\columnwidth}
		\centering
		\includegraphics[width=1.0\columnwidth]{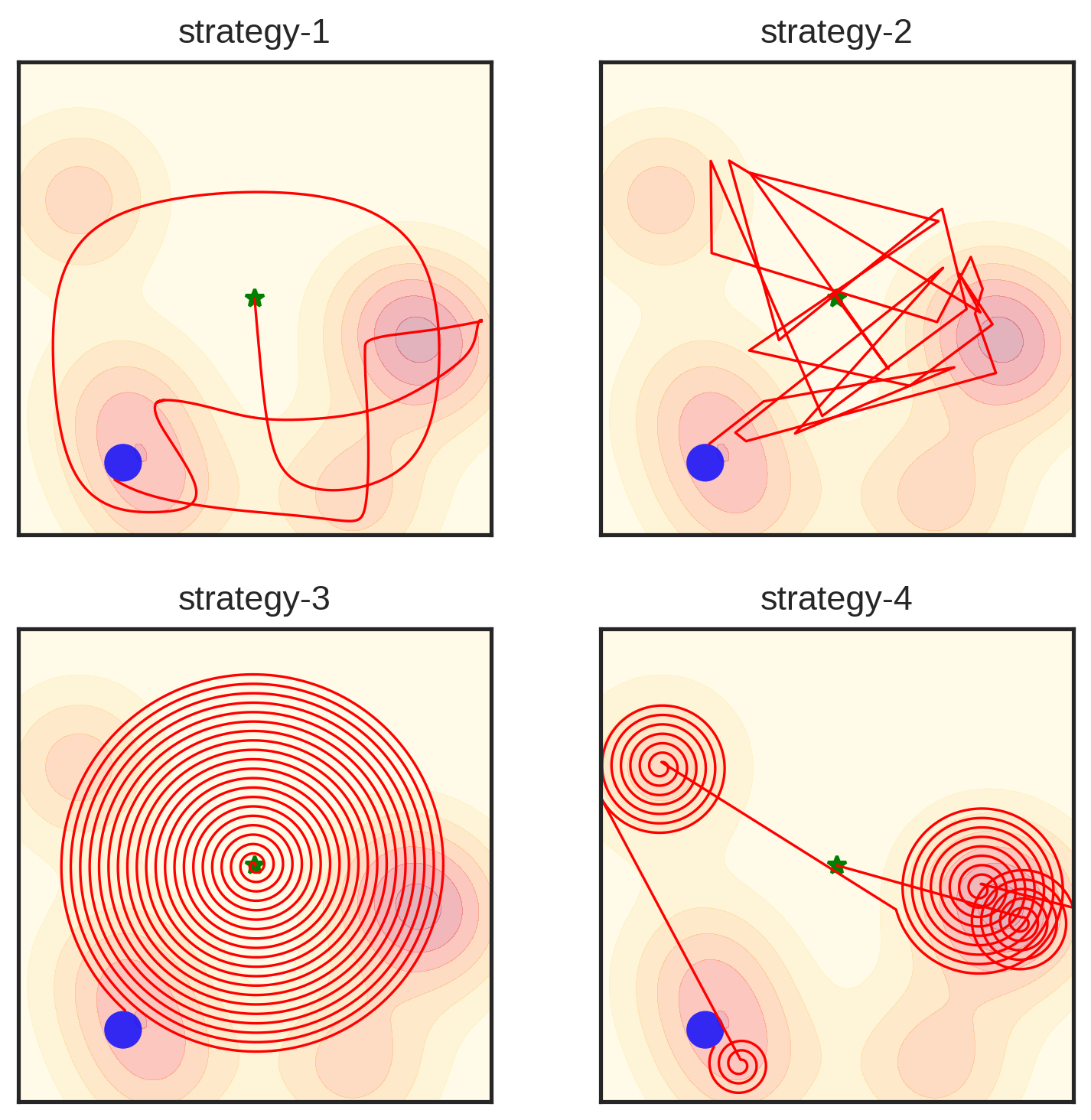}
		\label{fig:first_reach_time_2D}
	\end{subfigure}%
	\\
	\vspace{0.25cm}
	\begin{subfigure}{0.75\columnwidth}
		\centering
		\includegraphics[width=1.0\columnwidth]{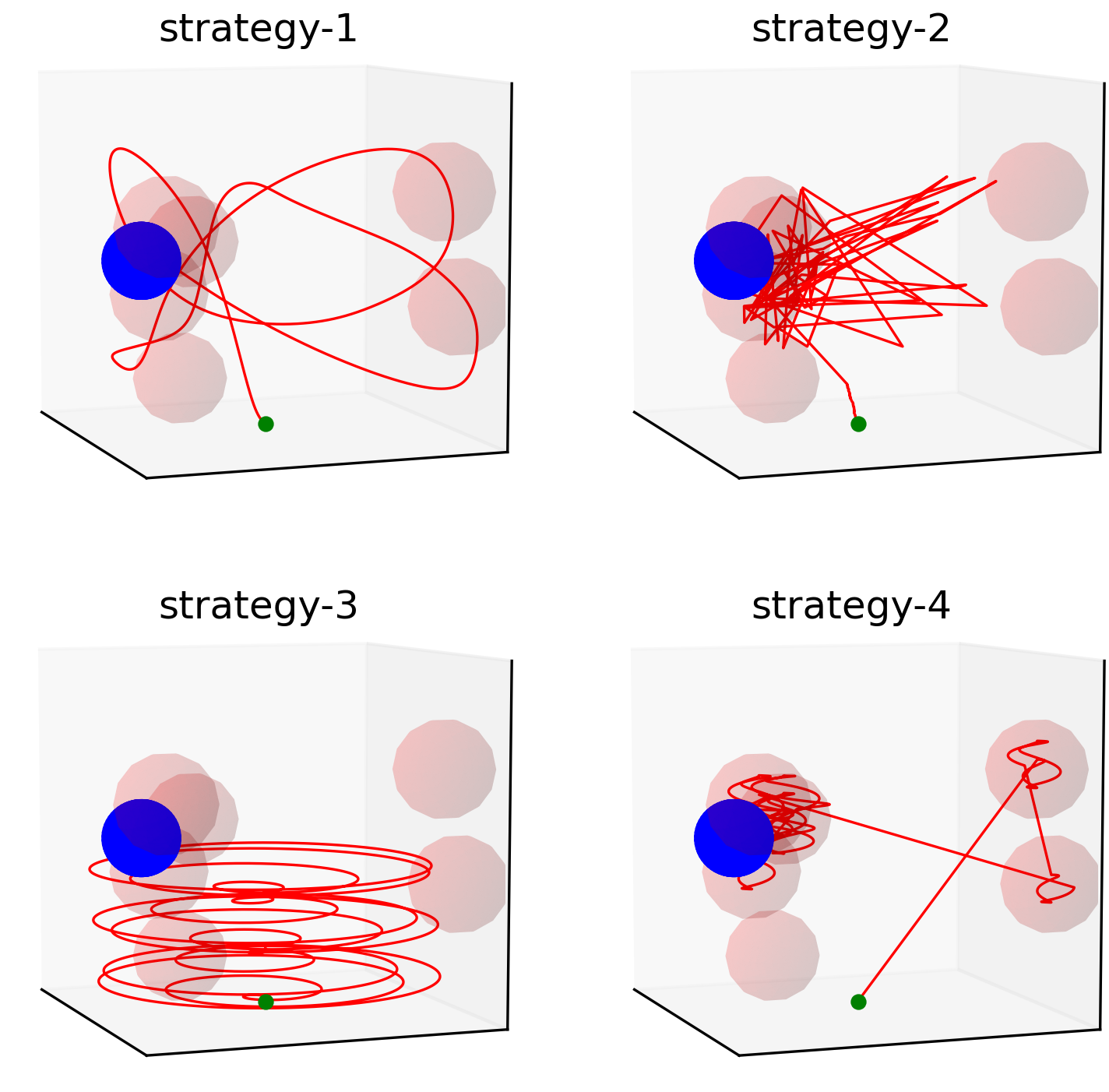}
		\label{fig:first_reach_time_3D}
	\end{subfigure}
	\caption{Example trajectories (red) of the four different exploration strategies to reach an target region (blue sphere) within a reference probability distribution (in this case a GMM) for $d=2$ (top) and $d=3$ (bottom). The green point indicates the initial state of the point-mass system. The goal is to reach this target region the information about which is known to the search strategies only through the reference probability distribution.} 
	
	\label{fig:first_reach_time}
\end{figure}

Fig.~\ref{fig:first_reach_time} shows an example of exploration behaviors for these different strategies. 

In strategy 2, we explore by tracking samples from the GMM sequentially. 
In this approach, 
the simulated system tracks GMM samples using a constant speed, with a new sample being computed every time the previous one is reached. Unlike ergodic exploration, such approaches based on sampling from the reference distribution are typically inefficient at handling distributed information \cite{miller}. 

In strategy 3, we use an Archimedean spiral 
for 2D and its cylindrical extension for 3D (see Fig.~\ref{fig:first_reach_time}). These are conventionally used in robotics as heuristic search strategies for uniform coverage in 2D and 3D search spaces (i.e. for uniform reference probability distributions) \cite{park2013intuitive_insertion} \cite{spiral2019}. Strategy 4 is similar to strategy 3 but uses spherical/ellipsoidal spiral trajectories that are customized to sweep the GMM search space. In this approach, the Gaussians are swept in sequence with  spherical/ellipsoidal spiral paths starting from the centers of the Gaussian and sweeping the area up to a given number of standard deviations before moving to the next component. These approaches suffer from the curse of dimensionality and perform poorly for $d>3$. Moreover, they 
require careful tuning of hyperparameters to generate efficient spiral paths. Furthermore, the resulting trajectories are  deterministic and do not consider the stochasticity of target detection. Namely, the trajectory generated by such sweeping pattern passes through a given point in the search space only once. If the measurement system fails to detect the target during its first pass, the strategy has no future possibilities for detection.

The first metric we use for comparison is the average time taken to reach the target region for the first time. Table \ref{tab: first_reach} shows the obtained results for 2D and 3D. We see that, on average, ergodic exploration reaches the target region faster due to its multiscale search behavior. Additionally, ergodic control has a 100\% success rate. Despite being slower, spiral search is equally successful, but this success comes at the cost of optimally choosing the spiral parameters. This is often cumbersome in practice, especially in higher dimensions and considering that the tolerance of the detection region is often not known with high certainty.

\begin{figure}[t!]
    \centering
     \includegraphics[width=1.0\linewidth]{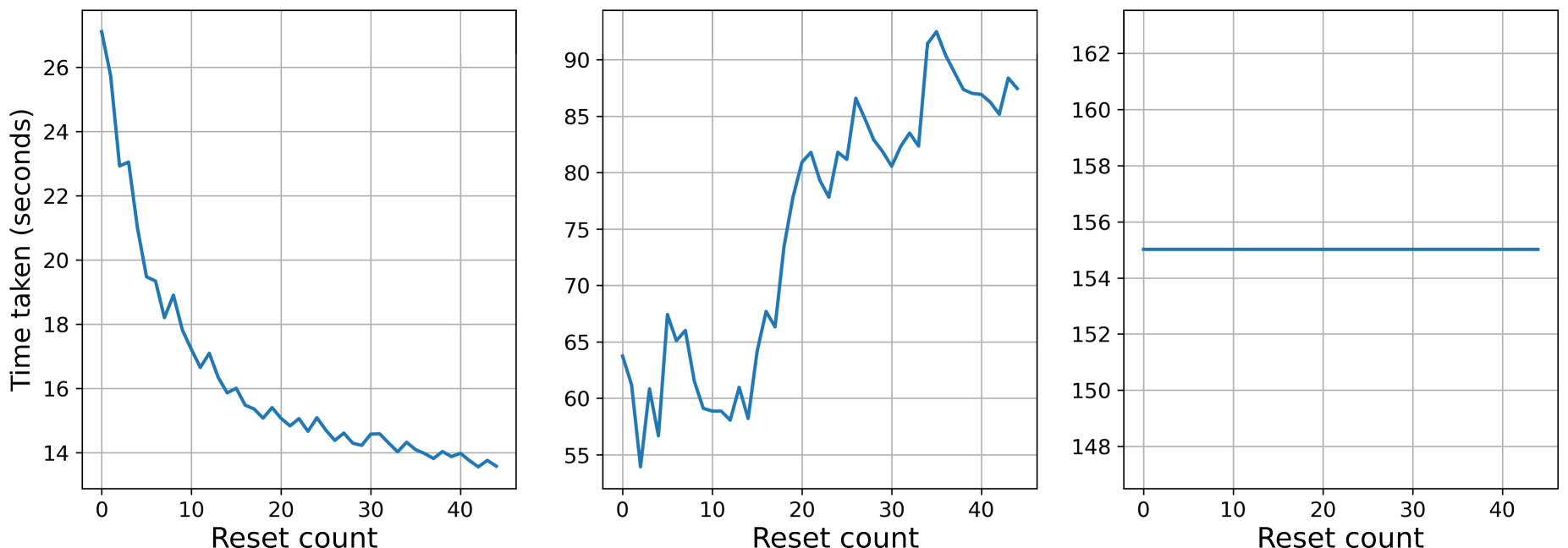}
     \caption{Cumulative average time to reach a  specified target region (spherical region with $1\%$ of the volume of $\Omega$ with $d=3$) with ergodic exploration (\textbf{left}), sampling-based strategy (\textbf{center}) and spiral movement (\textbf{right}). The x-axis represents the number of attempts to reach the target (reset count) and the y-axis represents the cumulative average of the time taken to reach the target at each number of attempts. Every time the target is reached, the system is re-initialized to the same initial state (starting a new attempt). The ergodic controller is aware of the lack of exploration inside the target region and tries to visit it more frequently. The cumulative average therefore converges to the time it takes to go from the initial state directly to the target at every re-initialization.}
     \label{fig:insertion_simulation}
\end{figure}


A desirable property for exploration strategies is that the system takes into account the already visited regions to cover the unvisited regions more often. In order to evaluate this property, we consider a second metric: the cumulative average time to reach the target region over several successful attempts. We define this as $\frac{T_c}{c}$, where $c$ is the number of successful attempts and $T_c$ is the cumulative time until successful attempt $c$. In this evaluation, as soon as the system reaches the boundary of the target region, we re-initialize it to the initial state and repeat this process for a fixed number of times. 
%
%
The results in Fig.~\ref{fig:insertion_simulation} show that, for ergodic control, the cumulative average decreases with the number of successful attempts and converges to a fixed value. This suggests that the ergodic exploration tries to visit the unexplored regions more frequently as the number of attempts increases. In this case, the value that the cumulative average converges to is the time it takes to go from the initial state directly to the target at every re-initialization. This is an essential feature for insertion tasks as the exploration inside the hole (representing successful insertion of the peg) is not easy due to obstacles (e.g., uncertainties and collision of the surface of the hole against the peg) and we need the exploration strategy to drive exploration towards the unexplored region as the time evolves. This is an inherent property of ergodic exploration that the other approaches lack. This property plays a crucial role to cope with the stochasticity involved either in the measurement systems for detection, or the dynamics of the process (e.g., insertion task). To exploit this feature in real robot experiments, it is necessary that the ergodic controller is implemented \textit{online} on the robot manipulator, i.e., that the controller knows the actual observed end-effector states. Our proposed algorithm for ergodic controller allows this online implementation on robot manipulators for $d=6$, which is demonstrated in the next section.

 In the insertion task, the target region for detection corresponds to a set of states of the peg that are necessary to be passed through for a successful insertion of the peg. This information is obtained from the reference probability distribution for exploration in the search space. Considering stochasticity is important for a search strategy to be useful in practice. In general, stochasticity may arise either from the measurement system (e.g., uncertainties in the location of the hole, grasp of the peg or manufacturing defects) and/or the dynamics of the system (e.g., stochasticity in the contact dynamics involved in the insertion). For example, during the insertion, the peg might be at the correct relative location to the hole according to the sensor system, but the insertion may still not be successful every time in that configuration due to the stochasticity of the process. We need the search strategy to explore more often in these target regions (correct configurations for insertion) to increase the likelihood of insertion. Ergodic control considers this stochasticity by driving the system to regions in the state space such that the amount of time it spends there is in proportion to the probability mass of that region. For more details on search strategies and their desired characteristics, see the seminal work of Koopman on the theory of search \cite{koopman1956theory_I,koopman1956theory_II,koopman1957theory}. Ergodic exploration satisfies the standards for optimal search behavior set by Koopman. Although strategies $3$ and $4$ do not satisfy these properties, we included them in our evaluation for completeness.
 
\begin{table}
	\renewcommand{\arraystretch}{1.3} 
	\centering
	\caption{Average time taken to reach a target region for different exploration strategies.}
		\begin{tabular}{ccccc}
			\toprule[0.12em]&
			{\multirow{2}{*}{\textbf{Strategy}}} & \multicolumn{2}{c}{\textbf{Success rate}} & \multirow{2}{*}{\textbf{Time taken}}\\
			\cmidrule{3-4}
			& & \# Trials & \# Success &  \textbf{(seconds)} \\
			\midrule
			\multirow{3}{*}{\textbf{2D}} 
			& Strategy 1 & $100$ &  $100$ & $\mathbf{66.9}$     \\
			& Strategy 2 & $100$ & $96$ & $106.8$  	\\
			& Strategy 3 & $100$ &  $100$ & $122.9$   \\
			& Strategy 4 & $100$ &  $98$ & $155.4$     	\\
			
			\cmidrule{2-5}			
			\multirow{3}{*}{\textbf{3D}} 
			& Strategy 1 & $100$ &  $100$ & $\mathbf{84.7}$   \\
			& Strategy 2 &  $100$ & $92$ & $141.4$         \\
			& Strategy 3  & $100$ & $98$ & $292.3$\\
			& Strategy 4 & $100$ & $95$ & $247.5$\\

			\bottomrule[0.12em]
		\end{tabular}
	\label{tab: first_reach}
	\vspace{-0.5cm}
\end{table}

\subsection{Experimental Setup for Peg-in-hole Task}
\begin{figure}[t!]
    \centering
     \includegraphics[width=0.5\linewidth]{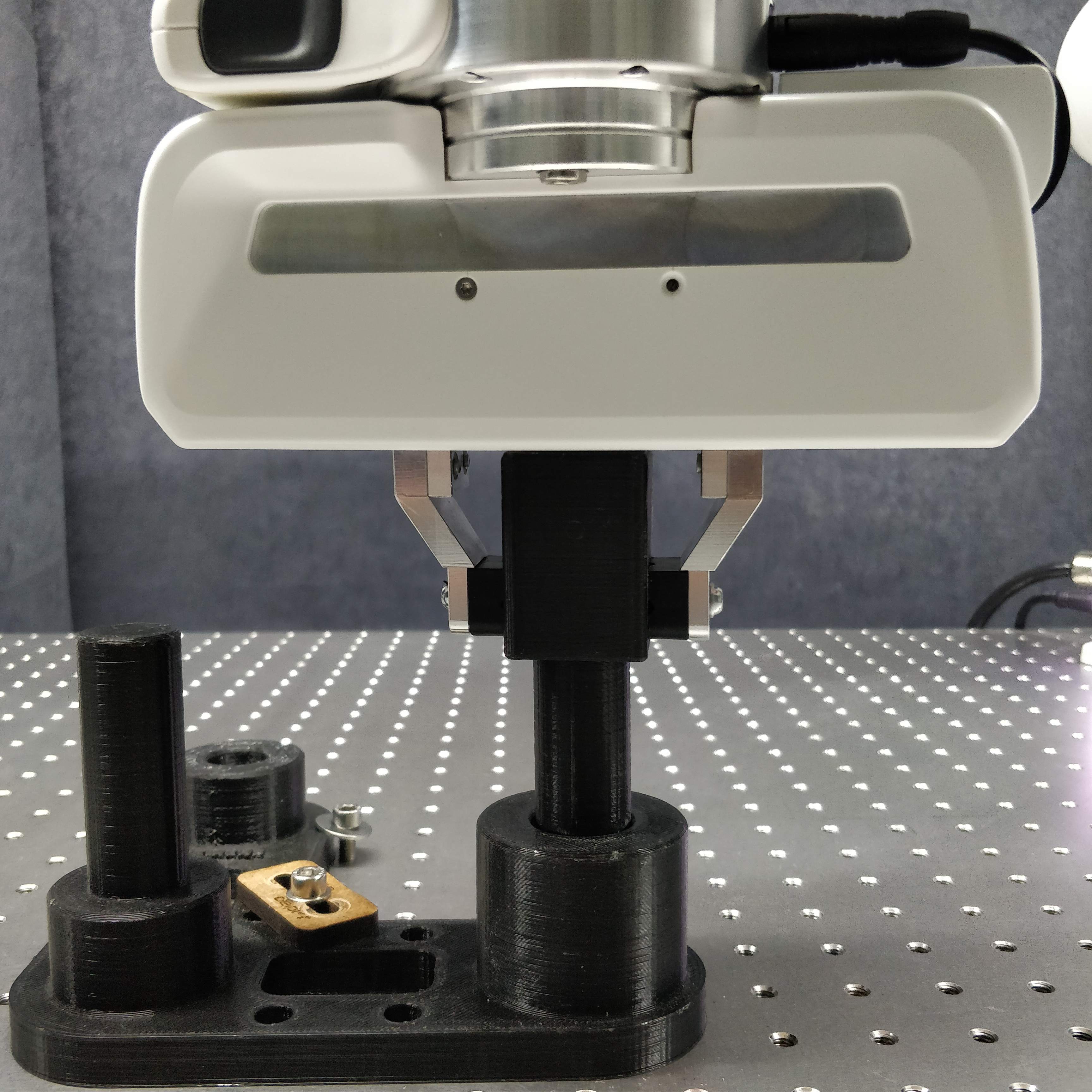}
     \caption{The Siemens gear benchmark used for evaluation of the peg-in-hole insertion task using ergodic control.}
     \label{fig:insertion_setup}
\end{figure}

We use a torque-controlled 7-axis Franka Emika Panda robot, with an insertion task based on the Siemens gear set benchmark (see Fig.~\ref{fig:insertion_setup})\footnote{\url{https://new.siemens.com/us/en/company/fairs-events/robot-learning.html}}, by using the $25.4\text{mm}$-diameter peg and the $26.29\text{mm}$-diameter receptacle, with the length of insertion of $47\text{mm}$.
We employed a Cartesian impedance controller to control the robot end-effector by computing a desired Cartesian wrench according to
\begin{equation*}
	\hat{\bm{f}} = \bm{K}_p \begin{bmatrix} \hat{\bm{p}}-\bm{p} \\ \mathrm{Log}\left(\hat{\bm{q}}*\bar{\bm{q}}\right) \end{bmatrix} - \bm{K}_d \begin{bmatrix} \dot{\bm{p}} \\ \bm{\omega} \end{bmatrix},
\end{equation*}
where $\hat{\bm{p}}\in\mathbb{R}^3$, $\hat{\bm{q}}\in\mathcal{S}^3$ are, respectively, the desired position and orientation of the end-effector (with $\hat{\bm{q}}$ obtained from \eqref{eq:expmap_tracking}), $\bm{p},\bm{q},\dot{\bm{p}},\bm{w}$ are the end-effector position, orientation, linear and angular velocity and $\bm{K}_p,\bm{K}_d$ are $6{\times}6$ diagonal stiffness and damping gains, experimentally set to $\bm{K}_p = \mathrm{diag}\left(500,500,500,160,160,160\right)$ and $\bm{K}_v = \mathrm{diag}\left(40,40,40,10,10,10\right)$. The symbol $*$ denotes the unit quaternion product and $\bar{\bm{q}}$ the conjugate of quaternion $\bm{q}$. $\mathrm{Log(\cdot)}$ is the logarithmic map defined in \eqref{eq:logmap_origin}. 

We obtain the desired robot joint torques with ${\hat{\bm{\tau}} = \bm{J}(\bm{\theta})^\trsp \hat{\bm{f}} + \bm{g}\left(\bm{\theta}\right) + \bm{h}(\bm{\theta},\dot{\bm{\theta}})}$, where $\bm{\theta},\dot{\bm{\theta}}\in\mathbb{R}^7$ are the robot joint positions and velocities and $\bm{J}\in\bm{R}^{6\times 7}$, $\bm{g}\in\mathbb{R}^7$, $\bm{h}\in\mathbb{R}^7$ are the Jacobian matrix of the end-effector, gravity and Coriolis terms. Note that the impedance gains were selected such that the robot remained compliant enough to safely interact with the environment during exploration, while still being able to track the ergodic trajectory.

\begin{figure}
    \centering
    \includegraphics[width=1.0\columnwidth]{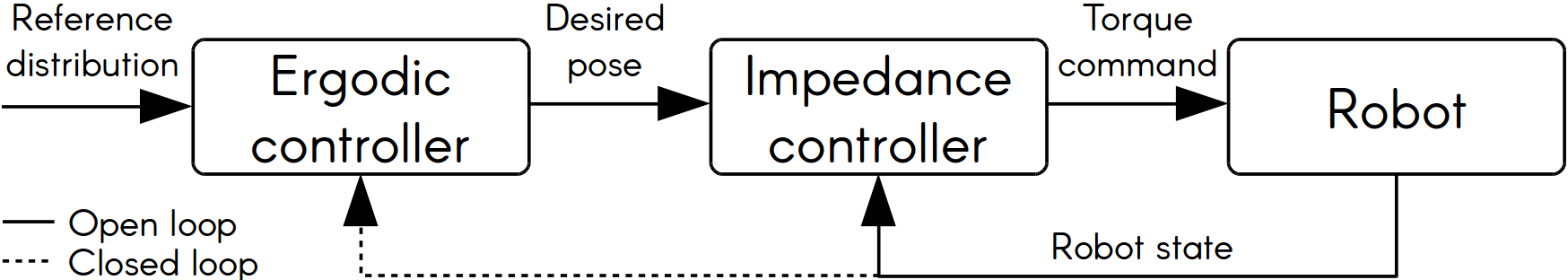}
    \caption{Control diagram of \textit{open} and \textit{closed} loop implementations. In the latter, the ergodic controller takes the robot state (end-effector pose) into account while computing the desired pose. In this way, the history of observed states is kept and new desired poses are computed accordingly.}
    \label{fig:ergodic_flow}
\end{figure}

We compare three different implementations of our approach. First, a \textit{closed loop} version where the ergodic controller runs in real-time on the robot as an online coverage problem (Fig. \ref{fig:ergodic_flow}). In that case, at every time step, we read the end-effector pose and feed it back to the ergodic controller, which computes the next pose to track based on the current state. Second, an \textit{open loop} version where the controller tracks a reference ergodic trajectory computed offline (as used in \cite{Ehlers19} for a lower dimensional state space). Lastly, GMM-sampling-based exploration as presented in Section \ref{why_ergodic_insertion}.

\begin{figure}%
	\begin{subfigure}{0.495\columnwidth}
		\centering
	   	\includegraphics[width=0.7\textwidth]{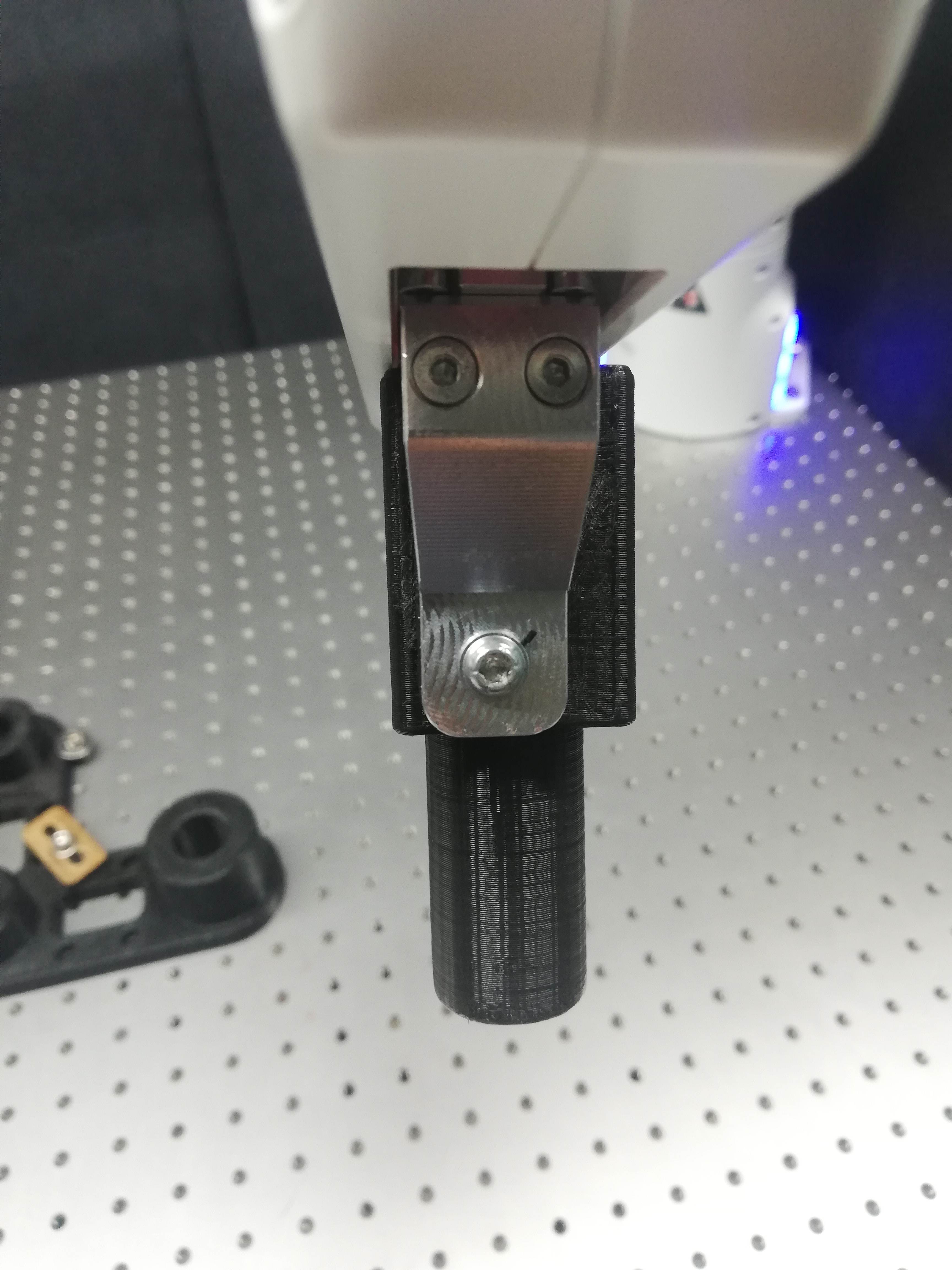}
   		\caption{\centering A grasp with an offset in position}
   		\label{fig:grasp_pos}
    \end{subfigure}  
   	\begin{subfigure}{0.495\columnwidth}
		\centering
   		\includegraphics[width=0.7\textwidth]{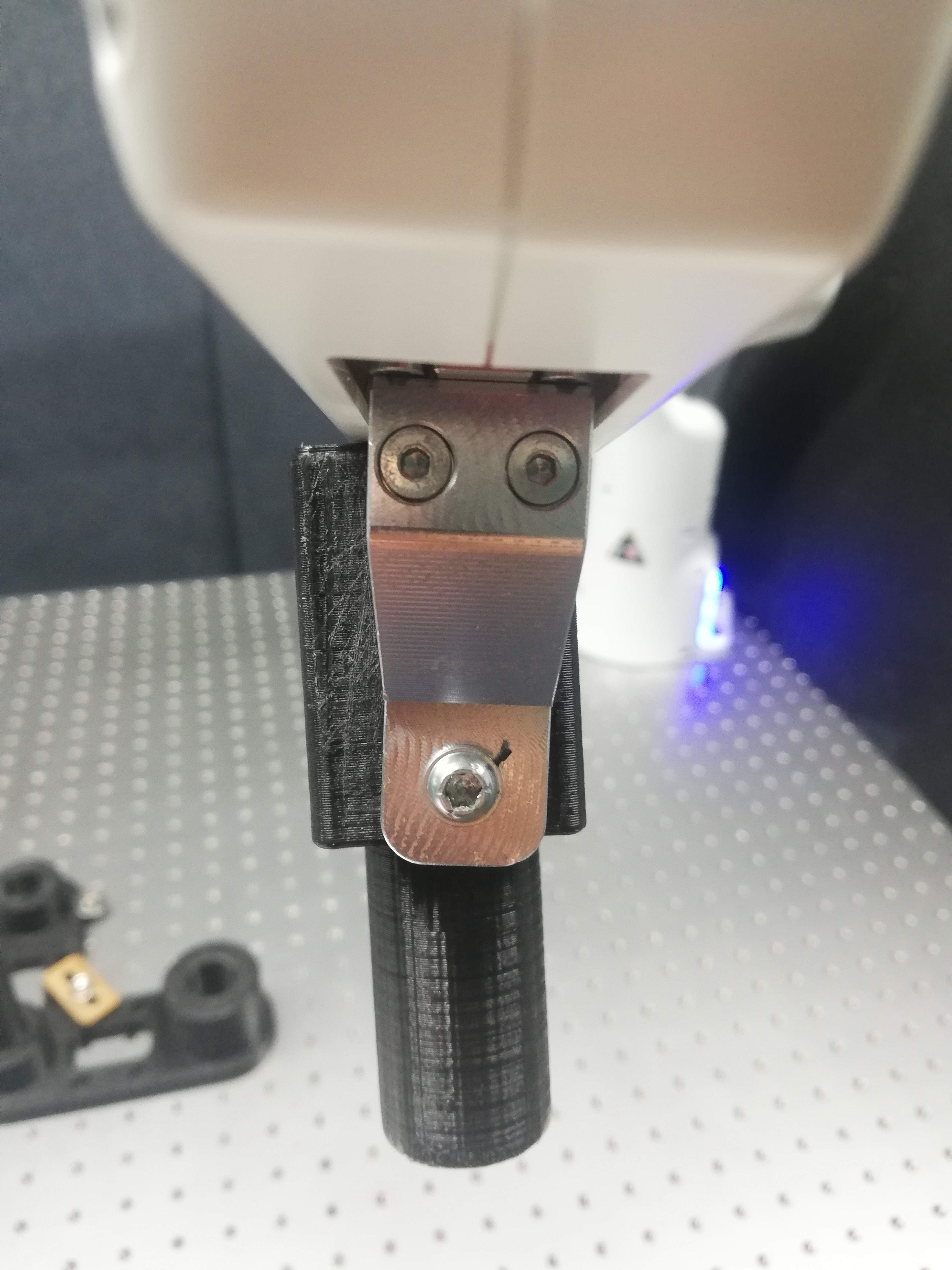}
   		\caption{\centering A grasp with an offset in orientation}
   		\label{fig:grasp_orient}
	\end{subfigure}   		
    \caption{Two instances of grasps typically appearing when testing the ergodic control for insertion. The demonstrations included different types of grasps to let the robot cope with this uncertainty during ergodic exploration.}
    \label{fig:example}%
\end{figure}

\subsection{Ergodic Controller Initialization}
In our setup, the location of the hole is fixed (but unknown to the robot) and the main source of uncertainty comes from the different possible grasps of the peg by the end-effector, see Fig.~\ref{fig:example}. We model these uncertainties using a probability density function that indicates the importance of spending time in each region of the state space of the robot end-effector. We use ergodic control to insert the peg under such uncertainties.

\begin{figure}[t!]
    \centering
     \includegraphics[width=0.9\linewidth]{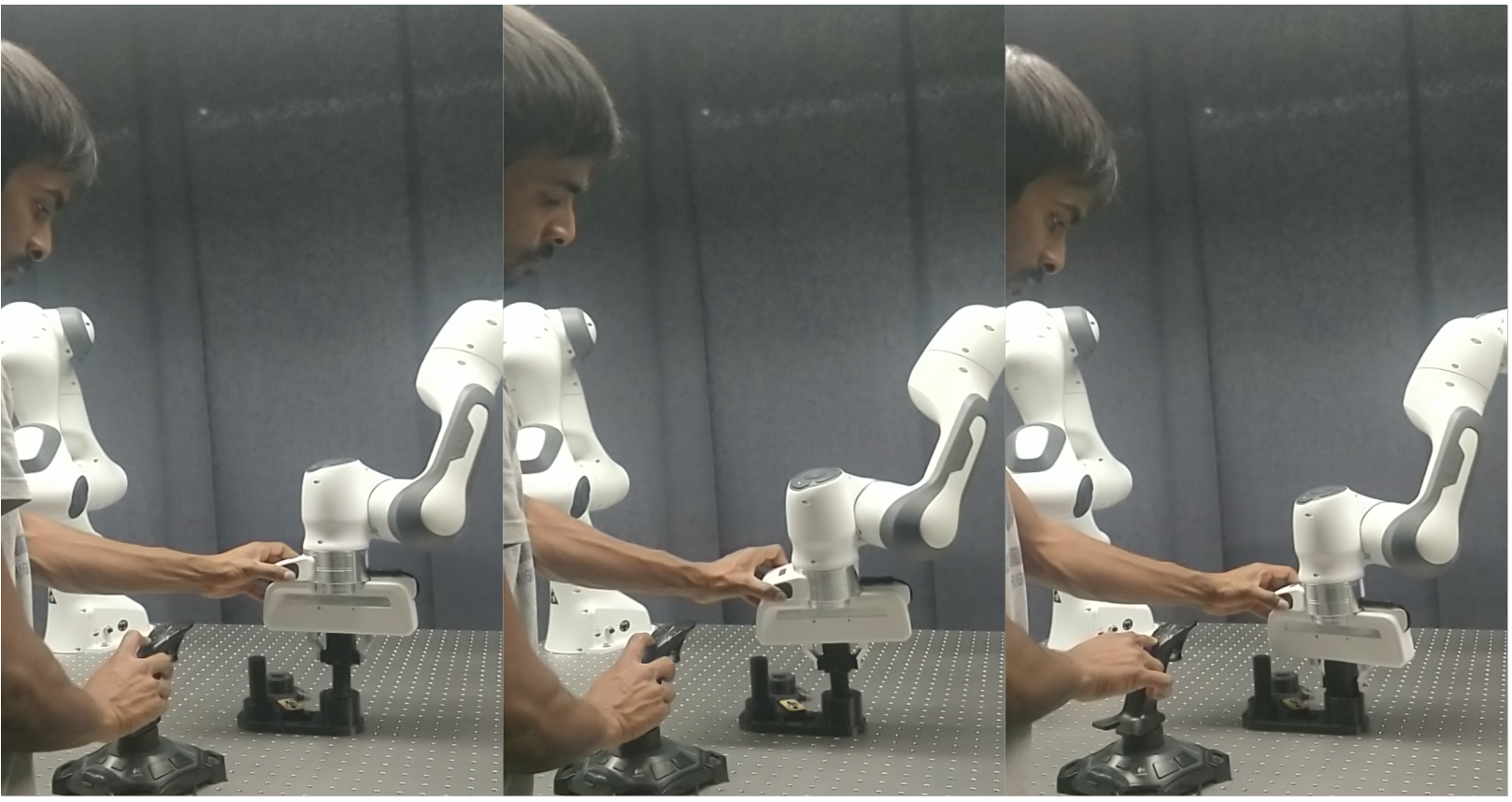}
     \caption{Human demonstration of peg-in-hole insertion task. Datapoints are collected for different grasps of the peg through kinesthetic teaching to show the regions in the vicinity of the receptacle to be used by the ergodic controller.}
     \label{fig:demo_insertion}
\end{figure}

To model the reference probability distribution for ergodic exploration of insertion, we collected $M=204$ data points using kinesthetic teaching, which corresponds to less than $2$ minutes of recording, see Fig.~\ref{fig:demo_insertion}.
Each datapoint corresponds to the position and orientation of the end-effector holding the peg. The datapoints in the vicinity of the location of the hole were recorded for successful insertions with different orientation and position offsets inherent to the grasps of the peg. To give higher importance to insertion, almost half the data points were taken from the states corresponding to the peg within the hole. A variation of about $15$mm in position of the grasp for each axis and about $\sim 10^\circ$ variation in the orientation of the grasp were demonstrated (see accompanying video of the experiment).

\begin{figure}
	\centering
	\begin{subfigure}{0.5\columnwidth}
		\centering
		\includegraphics[width=1.0\columnwidth]{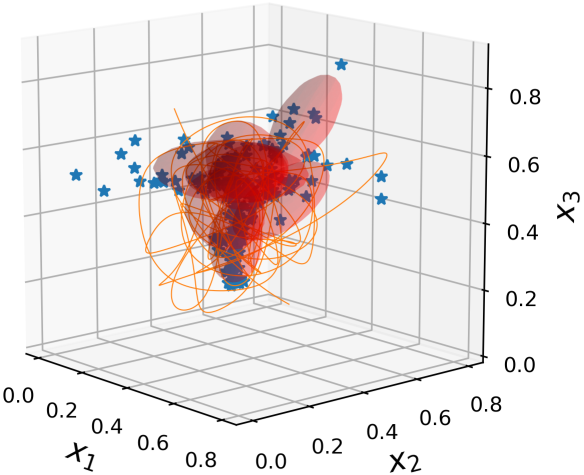}
		\caption{Position distribution}
		\label{fig:pos_GMM}
	\end{subfigure}%
	\begin{subfigure}{0.5\columnwidth}
		\centering
		\includegraphics[width=1.0\columnwidth]{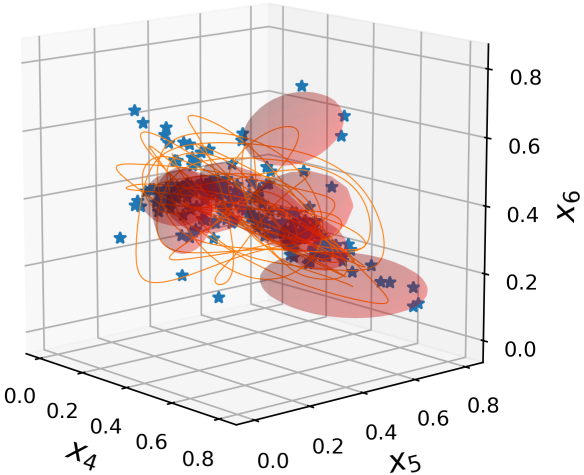}
		\caption{Orientation distribution}
		\label{fig:orient_GMM}
	\end{subfigure}
	\caption{Position and orientation marginals from the distribution used for full pose exploration (the figure does not show correlations between position and orientation). The 6D pose distribution is encoded as a GMM with 8 components (red ellipsoids) and full covariances, trained on a dataset of $M=204$  datapoints (blue points). The trajectory for the exploration (orange lines) is generated by ergodic control for the insertion task.}
	\label{fig:GMM}
\end{figure}

Once data are collected, we preprocess the end-effector orientations as described in Section \ref{sec:mani}. Subsequently, we concatenate the position data with the obtained 3D orientation representation into a 6D vector.  We then transform the data into the domain of ergodic control $\Omega$ with $L=1$ (ergodic space) using a bijective linear transformation. We model the data points in this transformed space using a Gaussian mixture model (GMM) with full covariances. We empirically selected 8 mixture components with a minimal isotropic covariance prior of $5\times 10^{-3}$. Figure \ref{fig:GMM} shows the obtained GMM for position and orientation (marginal distributions). The GMM is used as reference probability distribution $P(\bm{x})$ for ergodic control in $\Omega$. With $K=10$ and $N=10$, the computation of Fourier coefficients $\bm{\hat{\mathcal{W}}}$ for the reference probability distribution took less than 2 minutes. The closed-loop ergodic controller could be run at 100Hz ($dt=0.01$ in Algorithm \ref{algo2}) on the robot. The same settings are applied to the open-loop version of the insertion task. In this case, the trajectory is generated offline using ergodic control. It is then tracked using the above-mentioned settings of the impedance control. Figure \ref{fig:GMM} shows examples of generated ergodic trajectories. We allow a maximum of 40 seconds for insertion. Figure \ref{fig:erg_traj_robot_open_loop} shows trajectory  generated from ergodic control in one of the trials of the experiment. 

\subsection{Experimental Results}
\begin{figure}[t!]
    \centering
     \includegraphics[width=0.9\linewidth]{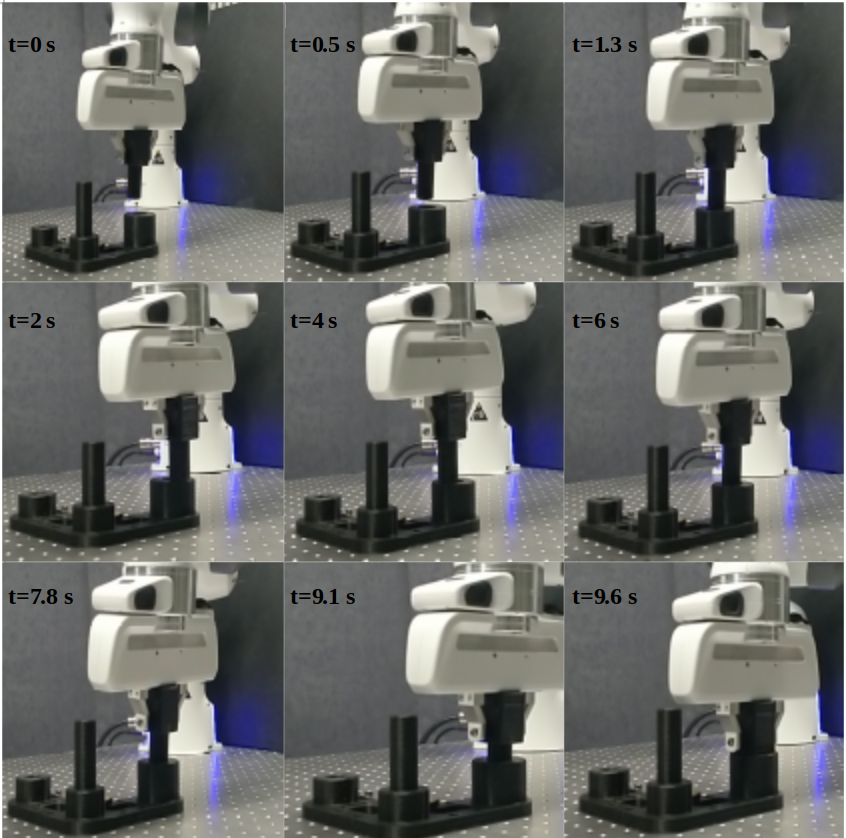}
     \caption{Snapshots of an insertion using ergodic control in closed loop.}
     \label{fig:insertion_snapshot}
\end{figure}

The obtained results, summarized in Table \ref{tab: success_table}, show that the \textit{closed loop} approach clearly outperforms the other approaches that do not consider the history of observed states during the exploration. Indeed, while using the former, the robot was able to successfully insert the peg in the hole in 20 out of the 20 trials, where 10 were performed using the grasp in Fig.~\ref{fig:grasp_pos} and the other 10 using the one in Fig.~\ref{fig:grasp_orient}. On average, the insertion using the \textit{closed loop} approach succeeded in $9.9$s with a standard deviation of $8.5$s. Figure \ref{fig:insertion_snapshot} shows snapshots of the insertion using ergodic exploration in the \textit{closed loop} setting. Note that the only required input was a set of demonstrations to show the robot the regions it should explore. This has proven important to deal with the uncertainty in the peg grasping pose. Note that interaction forces during exploration can also cause the grasp pose to change due to the limited gripping force of the robot (either from hardware and software limitations, or set on purpose to handle fragile objects). Our experimental results show that the closed loop ergodic strategy is also robust in these situations.

The open loop version succeeded in only $2$ out of $10$ trials and the naive approach of GMM-sampling-based exploration succeeded in none of the $5$ trials. This shows the importance of exploiting the history of the real observed end-effector poses to compute control commands. This is particularly important for tasks requiring contacts with the environment, where the commands need to be re-evaluated according to the history of poses retrieved by the impedance controller. Notably, this allows the use of low gains to remain compliant and enable safe contacts. In the closed loop approach, the algorithm is aware of the locations that were previously effectively visited, which is exploited to fulfill the insertion task in an online and robust fashion.\footnote{A video of the experiment is available at \url{https://sites.google.com/view/ergodic-exploration/}.}


\begin{table}
	\renewcommand{\arraystretch}{1.3} 
	\centering
	\caption{Performance of the peg-in-hole task for different variations of the grasps.}
		\begin{tabular}{cccc}
			\toprule[0.12em]
			\centering \multirow{2}{*}{\textbf{Strategy}} & \multicolumn{2}{c}{\textbf{Success rate}} & \multirow{2}{*}{\textbf{Time taken}}\\
			\cmidrule{2-3}
			& \# Trials & Successful Trials &  \textbf{(seconds)} \\
			\midrule
			Closed loop  & $20$ & $20$ & $9.9 \pm 8.5 $ \\
			Open loop  & $10$ &  $2$ & -  \\
			GMM-sampling & $5$ & $0$ & - \\
			\bottomrule[0.12em]
		\end{tabular}
	\label{tab: success_table}
\end{table}

\section{Conclusion and Further Work}
\label{sec:conclusions}
We proposed a solution to the multidimensional ergodic exploration problem with $d>3$, which was previously considered to be intractable for applications in robotic manipulation. The proposed approach relies on tensor train decompositions and is evaluated in simulated examples of target detection and a real robot experiment using a peg-in-hole insertion task. The obtained results show that ergodic control has a 100\% success rate in all tasks, succeeding faster than the competing approaches. In addition to the novelty of exploiting ergodic control in an online fashion for insertion tasks, the computational techniques we used demonstrate how algorithms in robotics, in general, can benefit from tensor methods to overcome the limitations in computational speed and storage requirements. To achieve this, tensor methods exploit the smoothness and symmetry of the functions underlying the problem. If the  underlying problem does not have low rank structure, there may not be significant saving in the storage and computational cost using tensor methods. However, ergodic control problem as formulated in SMC has such low rank structure by design.  

The main challenges of extending ergodic control to state space of more than 3 dimensions concern the computation of Fourier coefficients (preprocessing step) and the speed of the control loop (for online implementation). This article addressed both of these challenges using the properties of tensor train decomposition. We leveraged this improved ergodic control formulation to propose a sensorless strategy for peg-in-hole insertion tasks. We validated our approach with a compliant robot manipulator, where the 6D regions to explore are obtained from kinesthetic human demonstrations. Our experimental results show that by using our approach, the robot is capable of achieving challenging insertion tasks without force/torque sensors. 

By using the reproducible Siemens gears benchmark, insertion tasks with unknown gripping variations could be achieved in less than 10 seconds on average. This is, in part, due to the multi-scale exploration behavior that is inherent to the ergodic metric we employed. With such metric, the resulting controller first crudely explores the region of interest, and then progressively refines the search until insertion, efficiently exploiting information about the already covered regions. This is also due to the fast processing that we propose (through tensor train factorization), that allows ergodic control to be run in an online manner, even by considering distributions of full end-effector poses. Indeed, re-estimating the control commands on-the-fly allows the proposed ergodic control strategy to be used together with an impedance controller with low gains, which is important for tasks requiring contacts with the environment.

There is still room for improvement of the proposed approach. First, designing demonstrations for insertion tasks can be further optimized. In future work, we plan to study alternative demonstration strategies (including different distributions), which could speed up the insertion time using ergodic control. We will also apply the insertion tasks to different physical setups  by using different sensory modalities. We also plan to study applications of our algorithm to other applications in robotics, either requiring exploration, active sensing, or, more generally, for applications requiring multidimensional basis functions with a low rank structure.

\appendices
\section{Proof of Fourier Coefficients Decomposition}
\label{appdx_1}
A one-dimensional integral
\begin{equation*}
s = \int_{x=0}^l f(x) dx
\end{equation*}
can be computed numerically with the Gaussian-Quadrature (GQ) rule as
\begin{equation*}
s  = \sum_{j=1}^N \alpha_j f(x_j),
\end{equation*}
where $N$ represents the degree of approximation (specified by the user), $x_j$ represents the discretization points, and $\alpha_j$ are the corresponding weights. For any polynomial function of degree less than $2N-1$, the above summation gives exact result without any error in the integration. For a given $N$, $x_j$ and $\alpha_j$ can be computed and are readily available using software packages for scientific computing.


We need to evaluate the multidimensional integral \eqref{eq:multi_int} to find $\bm{\hat{\mathcal{W}}}_{\bm{k}}$. Let  $x_j$ be the discretization points and $\alpha_j$ the corresponding weights, with $j \in \{ 1,\ldots, N\}$, obtained from the GQ rule for a given $N$. Let $\mathcal{J} = \{\bm{j}=(j_1,\ldots, j_d):  j_i \in \{1, \dots, N\}\}$ be the index set. We can discretize the domain $\Omega$ at $\bm{x_j} = (x_{j_1},\ldots, x_{j_d})$, with $\bm{j} \in \mathcal{J}$. Let $\bm{\mathcal{P}}$ be the tensor formed by evaluating the reference distribution $P(\bm{x})$ at the discretization points, i.e., $\bm{\mathcal{P}_j} = P(\bm{x_j})$. 

Then, we can evaluate \eqref{eq:multi_int} using GQ as
\begin{align}
    \bm{\hat{\mathcal{W}}}_{\bm{k}} &=  \sum_{\bm{j} \in \mathcal{J}} \alpha_{j_1} \cdots \alpha_{j_d} P(\bm{x}_{\bm{j}})\bm{\Phi}_{\bm{k}}(\bm{x}_{\bm{j}})\nonumber\\
    &=  \sum_{{\bm{j} \in \mathcal{J}}} \alpha_{j_1} \cdots \alpha_{j_d} P({\bm{x_j}}) \phi_{i_1}(x_{j_1}) \cdots \phi_{i_d}(x_{j_d}), \; \forall \bm{k} \in \mathcal{K}.
    \label{gq_integral}
\end{align}

Discretizing $P(\bm{x})$ at the GQ points $\bm{x_j} = (x_{j_1},\ldots,x_{j_d})$ with $\bm{j} \in \mathcal{J}$, we can get the tensor $\bm{\mathcal{P}}$, with  $\bm{\mathcal{P}_j} = P(\bm{x_j})$. 

Consider a TT-representation of $ \bm{\mathcal{P}}$ given by the TT-cores $( \bm{\mathcal{P}}^1,\bm{\mathcal{P}}^2, \ldots, \bm{\mathcal{P}}^d )$, so that for $\bm{j}=(j_1,\ldots,j_d) \in \mathcal{J}$ we have
\begin{equation*}
    \bm{\mathcal{P}}_{\bm{j}} = \bm{\mathcal{P}}^1_{:,:,j_1} \bm{\mathcal{P}}^2_{:,:,j_2}\cdots \bm{\mathcal{P}}^d_{:,:,j_d}.
\end{equation*}

Substituting the above expression in \eqref{gq_integral} yields
\begin{align}
    \bm{\hat{\mathcal{W}}}_{\bm{k}} &=  \sum_{\bm{j}\in \mathcal{J}} \alpha_{j_1} \cdots \alpha_{j_d} \bm{\mathcal{P}}^1_{:,:,j_1}\cdots  \bm{\mathcal{P}}^d_{:,:,j_d}  \phi_{k_1}(x_{j_1}) \cdots \phi_{k_d}(x_{j_d}) \nonumber\\
    &=   \Big(\!\sum_{j_1=1}^N \alpha_{j_1}\bm{\mathcal{P}}^1_{:,:,j_1} \phi_{k_1}(x_{j_1})\!\Big) \!\cdots\! \Big(\!\sum_{j_d=1}^N \alpha_{j_d}\bm{\mathcal{P}}^d_{:,:,j_d} \phi_{k_d}(x_{j_d})\!\Big).
    \label{eq:tt_express_proof}
\end{align}
    
Also, we know that the TT-decomposition of $\bm{\hat{\mathcal{W}}}$ takes the form
\begin{equation}
    \bm{\hat{\mathcal{W}}}_{\bm{k}} = \bm{\hat{\mathcal{W}}}^1_{:,:,k_1} \cdots \bm{\hat{\mathcal{W}}}^d_{:,:,k_d}, \; \forall \bm{k} \in \mathcal{K}.
    \label{eq:tt_decomposed_appx}
\end{equation}
    
Comparing \eqref{eq:tt_express_proof} with \eqref{eq:tt_decomposed_appx}, we obtain an expression for the TT-cores of the TT-decomposition of $\bm{\hat{\mathcal{W}}}$ as
\begin{equation}
    \bm{\hat{\mathcal{W}}}^i_{:,:,k} = \sum_{j=1}^N \alpha_{j} \bm{\mathcal{P}}^i_{:,:,j} \phi_{k}(x_{j}),
    \begin{array}{l}
    \forall k \in (1,\ldots,K),\\
    \forall i \in (1,\ldots,d).
    \end{array}
    \label{W_expression_appdx}
\end{equation}



\ifCLASSOPTIONcaptionsoff
  \newpage
\fi

\bibliographystyle{IEEEtran}
\bibliography{main}

\begin{IEEEbiography}
    [{\includegraphics[width=1in,height=1.25in,clip,keepaspectratio]{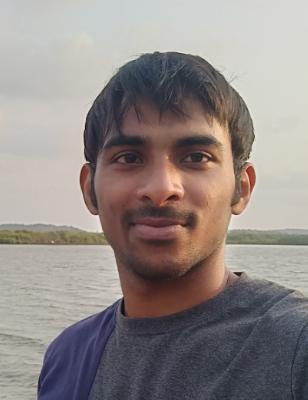}}]{Suhan Shetty}  received the M.E.~degree in mechanical engineering from the Indian Institute of Science, Bangalore, India. He is currently a Research Assistant with the Idiap Research Institute in Switzerland, working toward the Ph.D.~degree in electrical engineering with the \'Ecole Polytechnique Fed\'erale de Lausanne (EPFL). His Ph.D.~thesis research is focusing on low-rank approximation techniques in robot learning.
\end{IEEEbiography}

\begin{IEEEbiography}
    [{\includegraphics[width=1in,height=1.25in,clip,keepaspectratio]{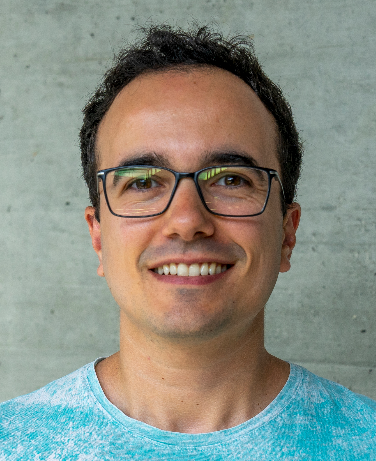}}]{Jo\~ao Silv\'erio} is a postdoctoral researcher at the Idiap Research Institute since July 2019. He received his M.Sc in Electrical and Computer Engineering (2011) from Instituto Superior T\'ecnico (Lisbon, Portugal) and Ph.D. in Robotics (2017) from the University of Genoa (Genoa, Italy) and the Italian Institute of Technology, where he was also a postdoctoral researcher until May 2019. He is interested in machine learning for robotics, particularly imitation learning and control. Webpage:
    \url{http://joaosilverio.eu}.
\end{IEEEbiography}

\begin{IEEEbiography}
    [{\includegraphics[width=1in,height=1.25in,clip,keepaspectratio]{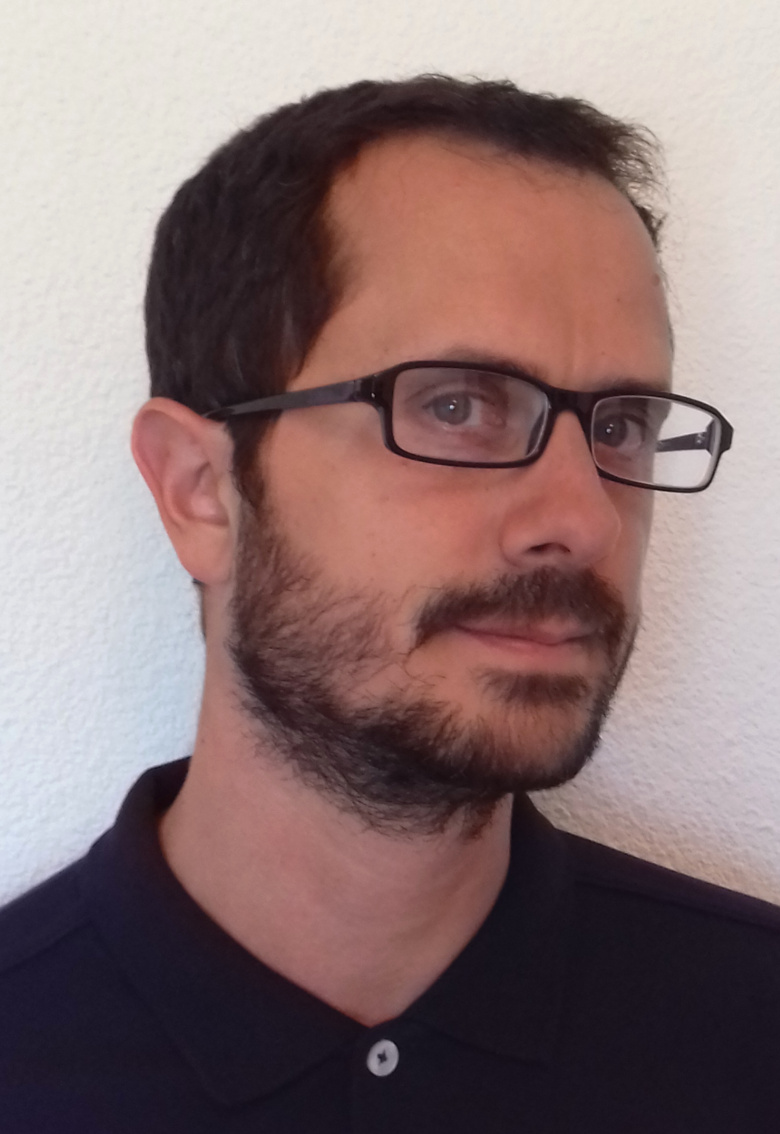}}]{Sylvain Calinon} received the Ph.D.~degree from the \'Ecole Polytechnique Fed\'erale de Lausanne (EPFL) in 2007. He is a Senior Researcher at the Idiap Research Institute, and a Lecturer at the EPFL. From 2009 to 2014, he was a Team Leader at the Italian Institute of Technology. From 2007 to 2009, he was a Postdoc at EPFL. His research interests cover robot learning, human-robot collaboration, Riemannian geometry and optimal control. Website: \url{https://calinon.ch}.
\end{IEEEbiography}

\end{document}